\pdfoutput=1
\documentclass{article}

\usepackage[table]{xcolor}

\usepackage{microtype}
\usepackage{graphicx}
\usepackage{booktabs} % for professional tables

\usepackage{hyperref}

\usepackage[accepted]{icml2024}
\usepackage{microtype}
\usepackage{graphicx}
\usepackage{booktabs} % for professional tables

\usepackage{mathtools}
\usepackage{algorithm}
\usepackage{tikz}
\usetikzlibrary{arrows,automata,positioning}
\usepackage{multirow} 
\usepackage{xspace}

\usepackage{comment}
\usepackage{amssymb} % define this before the line numbering.

\usepackage{dsfont}
\usepackage{tabulary}
\usepackage{tabularx}
\usepackage{makecell}
\usepackage{wrapfig}
\usepackage{enumitem}

\usepackage{dsfont}
\usepackage{tabulary}
\usepackage{tabularx}
\usepackage{makecell}
\usepackage{wrapfig}
\usepackage{subcaption}

\usepackage{amsmath}
\usepackage{amssymb}
\usepackage{mathtools}
\usepackage{amsthm}
\usepackage{array}

\usepackage[capitalize,noabbrev]{cleveref}

\theoremstyle{plain}

\theoremstyle{definition}

\theoremstyle{remark}

\usepackage[textsize=tiny]{todonotes}

\newcommand{\algabbr}{TVL\xspace}

\newcommand{\dataabbr}{HCT\xspace}

\definecolor{orange}{rgb}{1,0.5,0}
\definecolor{lightsalmonpink}{rgb}{1.0, 0.6, 0.6}
\definecolor{verylightsalmonpink}{rgb}{0.966, 0.805, 0.797}
\definecolor{lightblue}{rgb}{0.862, 0.906, 0.984}
\definecolor{lightyellow}{rgb}{1.0, 0.945, 0.797}
\definecolor{lightgreen}{rgb}{0.835, 0.91, 0.828}
\definecolor{lightpurple}{rgb}{0.879, 0.832, 0.902}

\newcommand{\ftsize}{\small}
\newcommand{\fsize}{small}
\usepackage[font=\fsize,labelfont=bf,skip=5pt]{caption}
\usepackage[export]{adjustbox}
\usepackage{wrapfig}

\usepackage{ragged2e}

\usepackage{listings}

\definecolor{codegreen}{rgb}{0,0.6,0}
\definecolor{codegray}{rgb}{0.5,0.5,0.5}
\definecolor{codepurple}{rgb}{0.58,0,0.82}
\definecolor{backcolour}{rgb}{0.95,0.95,0.92}

\lstdefinestyle{mystyle}{
    backgroundcolor=\color{backcolour},   
    commentstyle=\color{codegreen},
    keywordstyle=\color{magenta},
    numberstyle=\tiny\color{codegray},
    stringstyle=\color{codepurple},
    basicstyle=\ttfamily\footnotesize,
    breakatwhitespace=false,         
    breaklines=true,                 
    captionpos=b,                    
    keepspaces=true,                 
    numbers=left,                    
    numbersep=5pt,                  
    showspaces=false,                
    showstringspaces=false,
    showtabs=false,                  
    tabsize=2
}

\icmltitlerunning{A Touch, Vision, and Language Dataset for Multimodal Alignment}

\begin{document}

\twocolumn[
\icmltitle{A Touch, Vision, and Language Dataset for Multimodal Alignment}
\icmlsetsymbol{equal}{*}

\begin{icmlauthorlist}
\icmlauthor{Letian Fu}{sch}
\icmlauthor{Gaurav Datta}{equal,sch}
\icmlauthor{Huang Huang}{equal,sch}
\icmlauthor{William Chung-Ho Panitch}{equal,sch}
\icmlauthor{Jaimyn Drake}{equal,sch}
\icmlauthor{Joseph Ortiz}{comp}
\icmlauthor{Mustafa Mukadam}{comp}
\icmlauthor{Mike Lambeta}{comp}
\icmlauthor{Roberto Calandra}{sch2}
\icmlauthor{Ken Goldberg}{sch}
\end{icmlauthorlist}

\icmlaffiliation{sch}{UC Berkeley}
\icmlaffiliation{comp}{Meta AI}
\icmlaffiliation{sch2}{TU Dresden}

\icmlcorrespondingauthor{Letian Fu}{max.fu.letian@berkeley.edu}
\icmlkeywords{Machine Learning, ICML}

\vskip 0.3in
]
\printAffiliationsAndNotice{\icmlEqualContribution} % otherwise use the standard text.
\def\tabEncExp#1{
\begin{table}[#1]
\ssmall
\centering
\begin{tabular}{r|cc|cc|cc}
\toprule[1pt]
 & \multicolumn{2}{c|}{\textbf{Tactile-Text}} & \multicolumn{2}{c|}{\textbf{Tactile-Vision}} & \multicolumn{2}{c}{\textbf{Vision-Text}} \\
& Top-1 & Top-5 & Top-1 & Top-5 & Top-1 & Top-5 \\
\midrule[0.1pt]
CLIP & - & - & - & - & 28.4\% & 64.9\% \\
SSVTP & - & - & 0.2\% & 0.3\% & - & - \\
\algabbr & 36.7\% & 70.3\% & 79.5\% & 95.7\% & 28.4\% & 64.9\% \\
\bottomrule[1pt]
\end{tabular}
\caption{\textbf{Top-1 and Top-5 Accuracy} across different modality pairs. We find that the trained \algabbr encoder (ViT-Tiny) shows better tactile-language alignment than OpenCLIP's vision-language alignment, suggesting that vanilla CLIP may not capture tactile semantics well. 
Because SSVTP is trained on a subset of the \algabbr dataset, it does not generalize well across the entire \algabbr dataset, motivating the need to scale tactile-vision datasets.}
\label{tab:encoder}
\vspace{-20pt}
\end{table}
}

\def\figSplash#1{
    \begin{figure}[#1]
        \centering
        \vspace{0.08in}
        \includegraphics[width=1.0\linewidth]{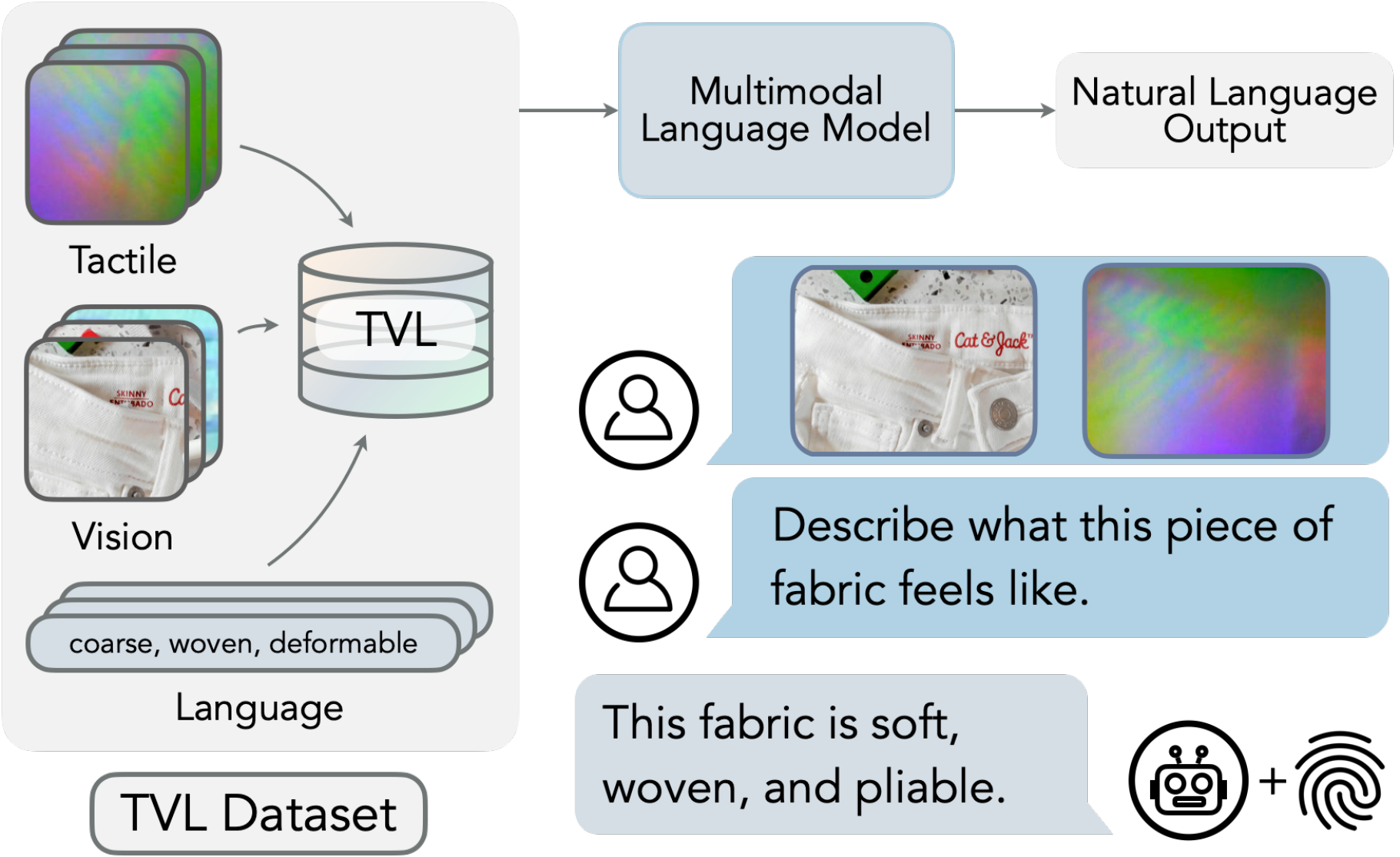}
        \vspace{-0.1in}
        \caption{
        \textbf{Can embodied agents integrate touch with vision and language?} 
        To the best of our knowledge, this work presents the first open-vocabulary tactile-vision-language dataset and we train 1) a vision-language aligned tactile encoder and 2) a tactile-vision-language model (TVLM) for describing tactile sensations.}
        \label{fig:splash}
         \vspace{-0.25in}
    \end{figure}
}

\def\figDevice#1{
    \begin{figure}[#1]
        \centering
        \vspace{0.08in}
        \includegraphics[width=1.0\linewidth]{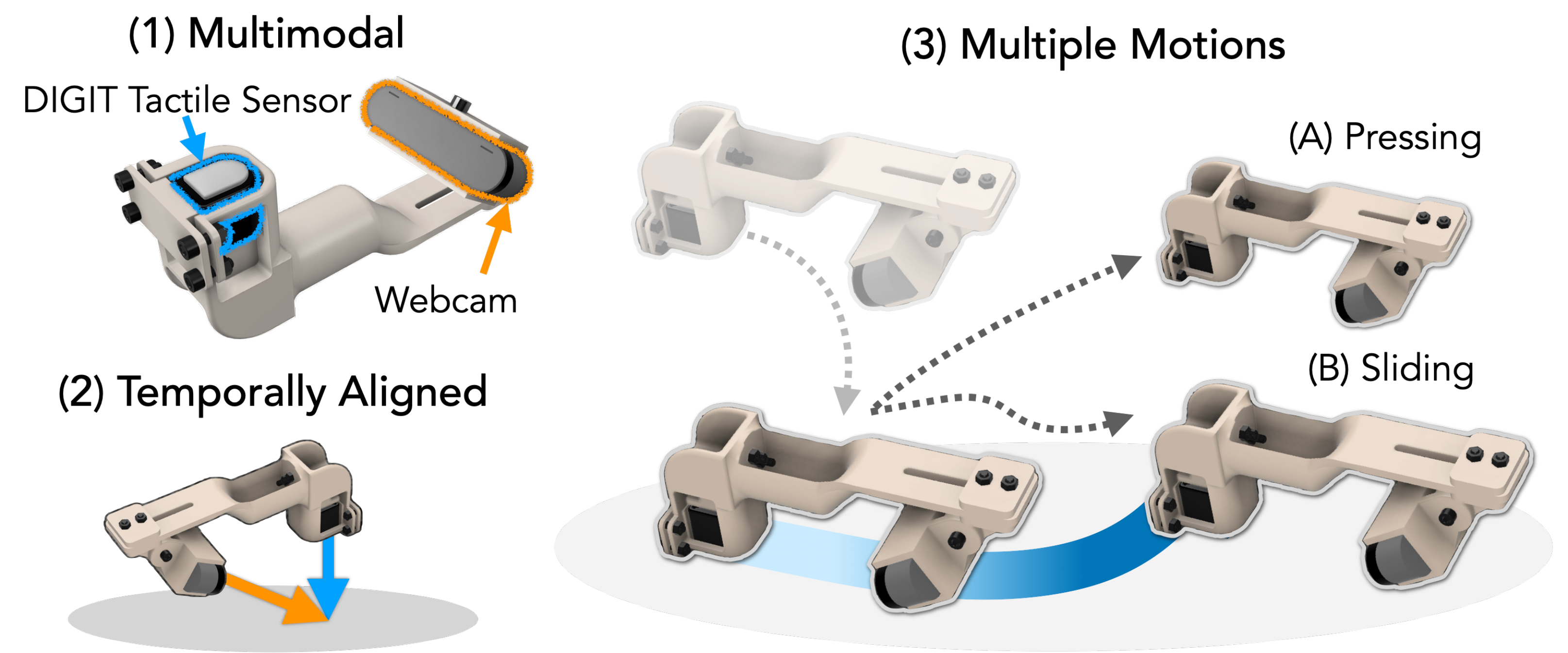}
        \vspace{-0.1in}
        \caption{
        (1) We designed a 3D printed data collection device using the DIGIT tactile sensor and a webcam to synchronously collect tactile and vision observations ``in-the-wild" (2). (3) We press and slide the device on surfaces and objects for data collection.}
        \label{fig:device}
         \vspace{-0.25in}
    \end{figure}
}

\def\figDataset#1{
    \begin{figure*}[#1]
        \centering
        \includegraphics[width=1.0\linewidth]{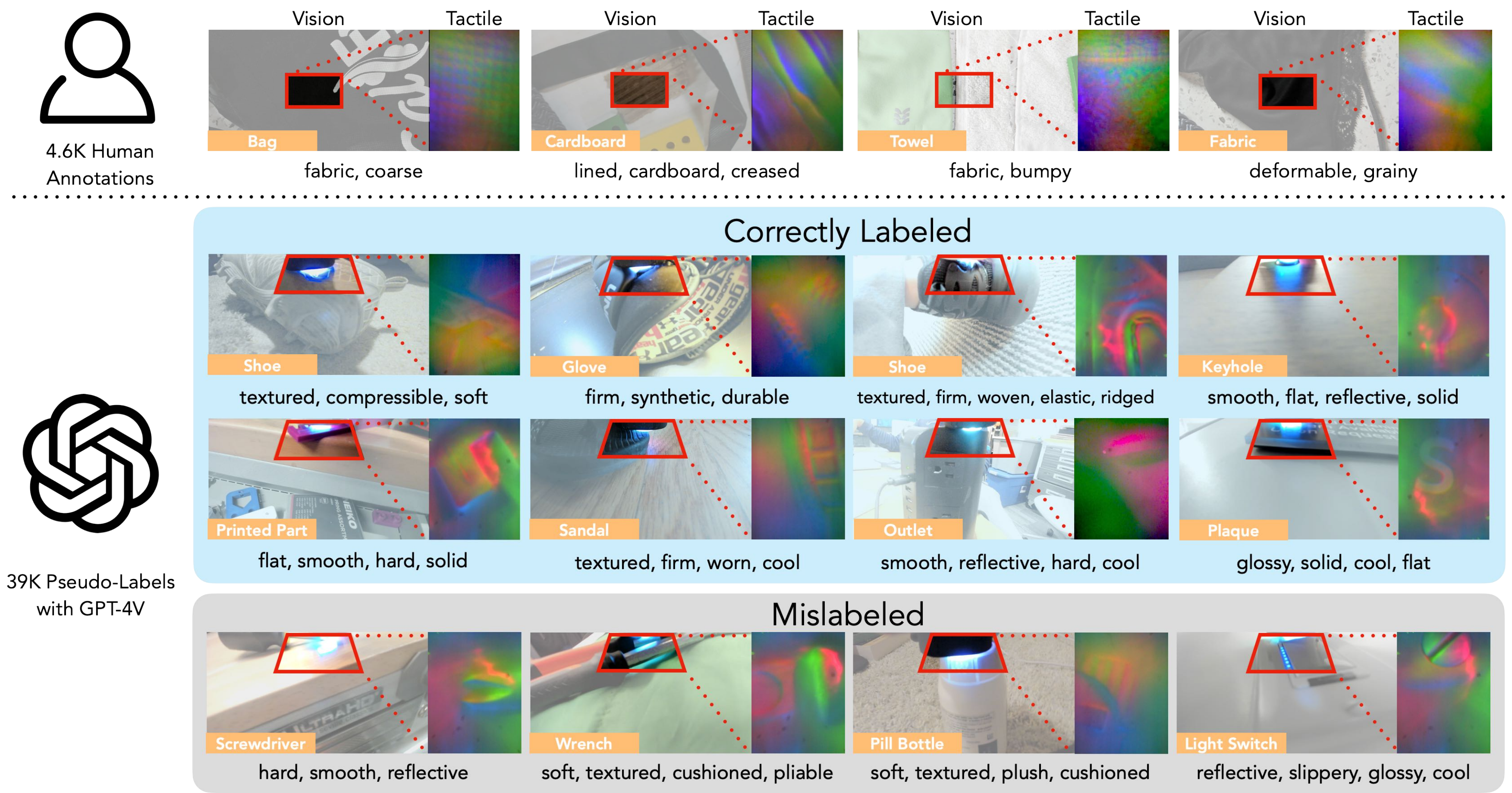}
        \vspace*{-0.1in}
        \caption{\textbf{\algabbr Dataset} starts by combining two datasets: SSVTP~\cite{kerr2023selfsupervised} (4,587 image-touch pairs) and \dataabbr (39,154 image-touch pairs), a new dataset we collected such that the visual observation and the tactile input are synchronously captured. 
        For the SSVTP dataset, we then manually label the data (examples shown in the first row). For the newly collected dataset, we prompt GPT-4V (see \cref{sec:appendix:pseudo}) to label the dataset (examples shown in rows 2-4). Note that GPT-4V will fail to provide correct tactile labels (row 4) when the contact patch is occluded by the sensor, or when there is not sufficient information to estimate the tactile sensation. In total, this results in a dataset containing 43,741 image-touch pairs with open-vocabulary language labels.}
        \label{fig:dataset}
        \vspace*{-0.15in}
    \end{figure*}
}

\def\figMethod#1{
    \begin{figure*}[#1]
        \centering
        \includegraphics[width=1.0\linewidth]{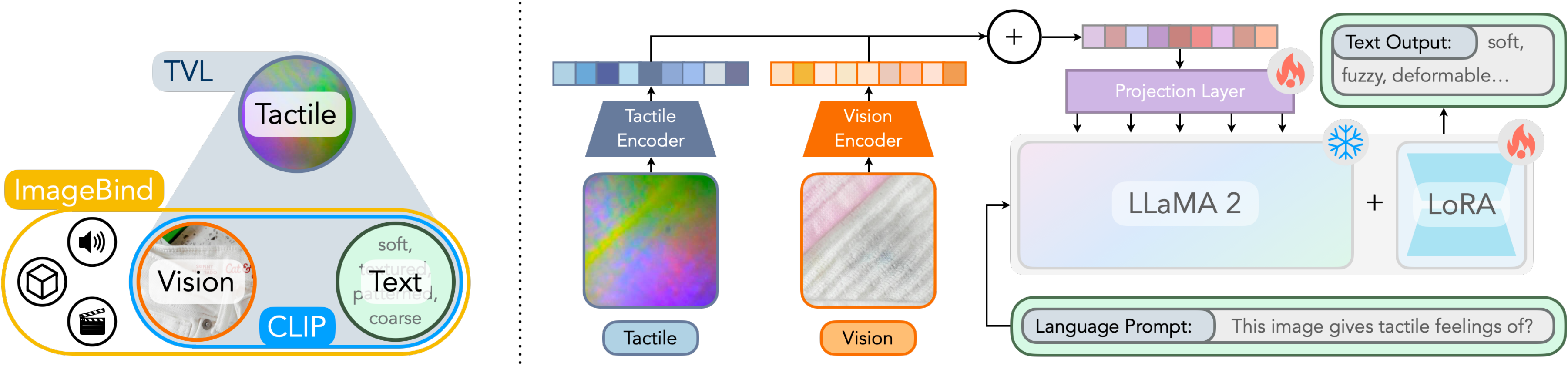}
        \vspace*{0in}
        \caption{\textbf{Method.} (Left) \algabbr is different from ImageBind~\cite{girdhar2023imagebind} as ImageBind only considers the loss between the vision modality and every other modality. \algabbr calculates loss between every pair of modalities, including that between the new modality (tactile) and language. Empirically, we show that including such loss can improve the model's capability to capture tactile semantics. (Right) Following~\citet{han2023imagebindllm}, we average the latent from the tactile and vision modality and finetune the language model.}
        \label{fig:model}
        \vspace*{-0.15in} 
    \end{figure*}
}

\def\figResults#1{
    \begin{figure*}[#1]
        \centering
        \vspace{0in}
        \includegraphics[width=1.0\linewidth]{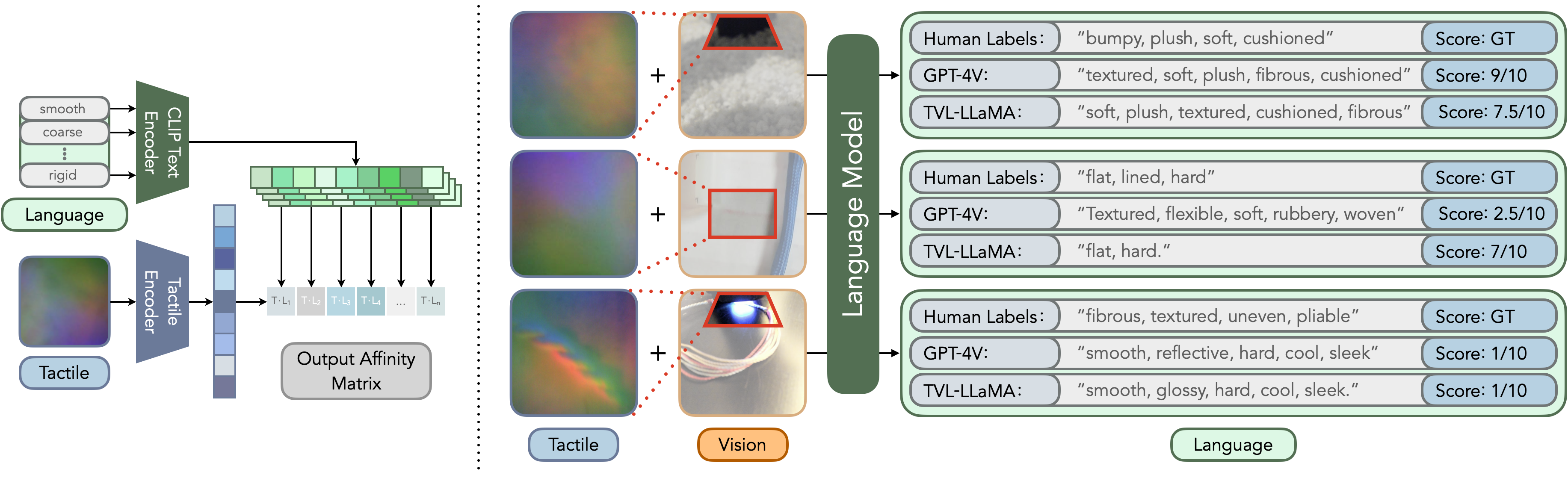}
        \vspace{-0.1in}
        \caption{\textit{Left:} We measure the cosine similarity between tactile and language on the entire test set containing 402 tactile, image, and language triplets. However, because different tactile observations may have synonymous language descriptions, in \ref{ssec:eval_metrics} we update top-1 and top-5 accuracy calculations to take this into account. \textit{Right:} GPT-4V and \algabbr-LLaMA generations with scores rated by GPT-4 based on the human labels. GPT-4V may be distracted by objects that are not in contact as it does not take tactile into account, and we empirically found there is no improvement when including tactile observation when prompting it because the observation is out-of-distribution. As TVL-LLaMA is trained on GPT-4V pseudo-labels, it suffers from the same failure mode.}
        \label{fig:results}
         \vspace{0in}
    \end{figure*}
}
\begin{abstract}
Touch is an important sensing modality for humans, but it has not yet been incorporated into a multimodal generative language model.
This is partially due to the difficulty of obtaining natural language labels for tactile data and the complexity of aligning tactile readings with both visual observations and language descriptions. 
As a step towards bridging that gap, this work introduces a new dataset of
44K in-the-wild vision-touch pairs, with English language labels annotated by humans (10\%) and textual pseudo-labels from GPT-4V (90\%).
We use this dataset to train a vision-language-aligned tactile encoder for open-vocabulary classification and a touch-vision-language (TVL) model for text generation using the trained encoder.
Results suggest that by incorporating touch, the \algabbr model improves (+29\% classification accuracy) touch-vision-language alignment over existing models trained on any pair of those modalities. 
Although only a small fraction of the dataset is human labeled, 
the \algabbr model demonstrates improved visual-tactile understanding over GPT-4V (+12\%) and open-source vision-language models (+32\%) on a new touch-vision understanding benchmark. 
Code and data: \url{https://tactile-vlm.github.io}.

\end{abstract}

\section{Introduction}
\label{sec:intro}

\figSplash{t!}
Almost all biological perception is inherently multimodal~\cite{bertelson2004psychology, turk2014multimodal, bruck2022cross}, enabling agents to reason and make decisions based on multiple streams of information. Recent research in artificial multimodal representation learning has explored linking modalities such as vision, language, audio, temperature, and robot actions~\cite{radford2021learning, girdhar2023imagebind, guzhov2021audioclip, rt22023arxiv, radosavovic2023robot}.
However, the tactile modality remains underexplored in multimodal understanding.
Touch enables humans to distinguish surface textures, object materials, dimensions, and contact forces~\cite{johansson2009coding, dahiya2009tactile, klatzky2003skin}. Tactile perception has also proven useful in robotic applications, particularly for contact-rich manipulation tasks
~\cite{digit, dahiya2009tactile, calandra2018more, yuan2017gelsight, dave2024multimodal, qi2023general}. 

Many works also explore visual tactile association, build cross-modal generators, and leverage cross-modal pertaining for material property, surface texture, and cloth classification on a closed set of vocabularies~\cite{yang2022touch, dave2024multimodal, li2013sensing, ojala2002multiresolution, kampouris2016multi, yuan2018active, kerr2023selfsupervised}. 

However, human tactile perception captures \emph{more} than tactile-visual associations; the tactile modality captures diverse semantic information and demonstrates deep integration with language~\cite{schmidt2019neuronal, speed2021crossmodal, miller2018verbal, ajbarnett_2023}. 
One major obstacle to the integration of touch and language is the scarcity of diverse data. While recent work has collected both datasets of paired tactile and visual observations and human-labeled datasets for tactile-based texture or material classification, we are not aware of any tactile dataset that contains open vocabulary language labels. Therefore, we develop a custom hand-held device (\cref{fig:device}) for synchronized ``in-the-wild'' touch-vision data collection, outside of a controlled laboratory setting. This setup allows us to capture \textit{close-up} visual observations and tactile readings while pressing and sliding on various foreground surfaces and objects with diverse backgrounds. 
\figDevice{t}
Another challenge is that human labeling can be costly and language descriptions of tactile experiences are subjective and vary between individuals. To address these challenges, we draw inspiration from prior works on training large language models (LLMs) and vision language models (VLMs)~\cite{alpaca, selfinstruct, liu2023llava, chen2023sharegpt4v}, which demonstrate vision language understanding by training on data synthesized by themselves or existing LLMs. We generate tactile descriptions from visual observations using an off-the-shelf LLM (GPT-4V~\cite{openai2023gpt4}) and hypothesize that it can serve as an effective captioner to mitigate the scarcity of labeled tactile-language data.

In this work, we present the \textbf{T}ouch-\textbf{V}ision-\textbf{L}anguage (\textbf{\algabbr}) dataset, a novel dataset 
consisting of 44K paired vision-tactile observations, where 10\% of the data are annotated by humans while the rest are labeled by GPT-4V.
Instead of binding all modalities to vision~\cite{girdhar2023imagebind}, we train a tactile encoder on this dataset by performing pairwise contrastive learning among all three modalities. 
We leverage existing vision and language encoders from OpenCLIP~\cite{ilharco_gabriel_2021_5143773} to train a tactile encoder that is aligned with both the textual and visual modalities. We evaluate alignment using the encoder's capability for touch-vision and touch-language classification.

Leveraging the dataset and the trained tactile encoder, we subsequently finetune LLaMA2~7B~\cite{touvron2023llama} to generate textual descriptions of tactile images based on visual and tactile observations (\cref{fig:splash}). To evaluate this model, we propose a Touch-Vision-Language Benchmark in which we query multimodal models to generate tactile descriptions and use an LLM to rate their consistency with ground truth human annotations. 

The proposed touch-vision-language model, trained on only a small amount of human-labeled data, demonstrates statistically significant improvement in performance on the TVL Benchmark when compared to open-source VLMs (+32\% improvement) and GPT-4V (+12\% improvement), the label-generating model. 

This paper makes the following contributions:
\begin{enumerate}[align=right,itemindent=2em,labelsep=2pt,labelwidth=1em,leftmargin=0pt,nosep]
    \item \textbf{TVL}, a new dataset containing 
    44K paired tactile-visual observations annotated with either human or VLM generated tactile descriptions, addressing the shortage of language-annotated tactile data;
    \item \textbf{A Vision-and-Language-Aligned Tactile Encoder} trained on the \algabbr dataset via pairwise contrastive learning between all three modalities and \textbf{a Touch-Vision-Language Model}, a multimodal model capable of generating tactile descriptions from both visual and tactile inputs;
    \item Experiments on the TVL Benchmark suggesting that a mix of human annotations and VLM pseudo-labels improves model performance in touch-vision-language understanding, surpassing existing VLMs by at least 12\%.
\end{enumerate}

\section{Related Work}
\label{sec:rw}

\subsection{Learning Multimodal Encoders}
Pretraining multi-modal encoders is a necessary step towards multi-task learning, as it can naturally structure the latent space to perform zero-shot cross-modal reasoning.
CLIP~\cite{radford2021learning, ilharco_gabriel_2021_5143773} is among the first to utilize internet-scale data to perform contrastive pretraining to learn a joint embedding space between vision and text. \citet{guzhov2021audioclip} and \citet{zhang2021pointclip, guo2023pointbind} extend CLIP to include audio and point clouds. ImageBind~\cite{girdhar2023imagebind} contrastively trains encoders for six modalities using only image-paired data. Many works also explored masking as an alternative strategy for multimodal pretraining~\cite{Bachmann2022, li2023scaling, geng2022multimodal}. In this work, we align the tactile modality with the CLIP latent space to capture its relationship with image observations and natural language descriptions of human tactility. 

\subsection{Tactile Perception}
Integrating tactile sensation with vision, inspired by the concurrent use of sight and touch in human perception~\cite{bresciani2006vision, ittyerah2007memory, jones2005comparison, camponogara2021integration, stone2015contributions}, is an active area of research in both robotics and embodied AI~\cite{goldberg1984active, pacchierotti2017}. 
Work in this field is facilitated by low-cost, vision-based tactile sensors
~\cite{chorley2009development, fingervision, yuan2017gelsight, digit, sferrazza2019design, shimonomura2019tactile}. %
Several recent works find that leveraging a combination of vision and touch helps with force and sensor pose estimation~\cite{suresh2022midastouch}, cross-modal image generation and prediction~\cite{higuera2023learning, zhong2022touching, yang2022touch, li2019connecting}, dexterous manipulation~\cite{calandra2018more, fu2023safe, 10185075, chen2022visuotactile, qi2023general, kerr2023selfsupervised}, and have produced datasets that include tactile, vision, and audio data~\cite{gao2021objectfolder, Gao_2022_CVPR}. 

Many works study the use of  tactile sensing for classifying surface textures, object material, and clothes. \citet{li2013sensing} 
classify 40 material properties from tactile observations using a non-learning-based texture classification method~\cite{ojala2002multiresolution}; subsequent works use learning-based methods for garment classification~\cite{kampouris2016multi, yuan2018active}.
By collecting data ``in-the-wild", \citet{yang2022touch} expanded the tactile observation diversity and trained a material classifier. 
All of these works use closed-vocabulary \textit{human} annotations of the entire dataset, whereas we use a vision-language model to label a dataset collected ``in-the-wild,"
and test on open-vocabulary tasks. Concurrent with this work, \citet{yang2024binding} binds touch to the vision modality, conducts open-vocabulary classification across tactile, vision, and language modalities, and aligns tactile inputs with language models for text generation without finetuning ImageBind-LLM~\cite{han2023imagebindllm}. 

\figDataset{t!}

\subsection{Multimodal Alignment in LLMs}
Pretrained multimodal encoders, when aligned with language models, enable language models to reason with non-text modalities.
Based on the capabilities of Large Language Models (LLMs), Unified-IO 2~\cite{lu2023unifiedio}, Generalist Agent~\cite{reed2022generalist}, Robot Transformer 2~\cite{rt22023arxiv}, and PaLM-E~\cite{driess2023palme} end-to-end finetune language models with internet and visual data from multiple domains. Recent work attempts to make alignment faster and more parameter efficient~\cite{zhu2023minigpt, moon2023anymal, dai2023instructblip, lin2023sphinx, chen2023shikra, cai2023making, bai2023qwenvl, hu2022lora}. Analogous to how open source language models train on GPT generated data~\cite{alpaca}, many vision-language models~\cite{liu2023llava, liu2023improvedllava, zhang2023llamaadapter, gao2023llamaadapterv2, chen2023sharegpt4v} finetune the model on language-image instruction-following data generated by GPT-4~\cite{openai2023gpt4} and show general visual reasoning capabilities. ImageBind-LLM~\cite{han2023imagebindllm} and PandaGPT~\cite{su2023pandagpt} introduce multimodal reasoning capability using ImageBind encoders. More recent work aligns pretrained LLMs, encoders, and decoders to finetune a model that can understand and generate multimodal data~\cite{wu2023nextgpt, tang2023codi2, sun2023generative}. Similar to Imagebind-LLM, this work aligns the multimodal encoder with a pretrained LLaMA-2~\cite{touvron2023llama}. 

\subsection{Training from Pseudo-labels}
The effectiveness of supervised learning is often limited by the availability of labeled data. Teacher models trained on a small set of labeled data can provide an inexpensive source of supervision in the form of pseudo-labels. A student model then learns from pseudo-labels generated by the teacher model on a large volume of unlabeled data~\cite{sohn2020fixmatch, lee2013pseudo, wang2022debiased, rosenberg2005semi, mclachlan1975iterative}. While previous works leverage training teacher models on labeled datasets, recent works in both vision and language literature leverage large-scale pretrained models. CutLER~\cite{wang2023cut} uses DINO~\cite{caron2021emerging} features to generate bounding boxes, enabling unsupervised training of object detection and segmentation models. InstructPix2Pix and InstructNeRF2NeRF~\cite{brooks2023instructpix2pix, instructnerf2023} use GPT~\cite{brown2020language} and Stable Diffusion~\cite{rombach2022highresolution} to generate a dataset of image editing examples and subsequently train a diffusion model based on these examples. Recent LLMs and VLMs~\cite{selfinstruct, alpaca, liu2023llava, liu2023improvedllava} are trained using pseudo-labels generated by GPT models~\cite{brown2020language, openai2023gpt4}. 
However, in these works the teacher and student models share the same input and output modalities. Similar to the framework proposed by \citet{burnel2023lesslabels}, we use a vision-only multi-modal model to generate textual labels from vision data, which in turn to match with tactile data to train the language-aligned tactile encoder and the TVL model. The teacher we use (GPT-4V) is more general than a specialist model trained on only the student task.

\section{\algabbr Dataset}
\label{sec:dataset}
The \algabbr Dataset (examples in \cref{fig:dataset}) contains paired tactile and vision observations labeled with tactile sensations in natural language. Here we describe the hardware and procedures used for data collection, cleaning, and labeling.

\subsection{Data Collection}
\label{ssec:dataset_background}
TVL uses vision data from a Logitech BRIO webcam and tactile data from DIGIT, a low-cost, compact, and open-source tactile sensor that provides high-resolution tactile observations in the form of RGB images of an internal deformable surface~\cite{digit}. The raw vision-tactile dataset amalgamates two distinct subsets: 1) the \textbf{S}elf-\textbf{S}upervised \textbf{V}isuo-\textbf{T}actile \textbf{P}retraining (SSVTP)~\cite{kerr2023selfsupervised} dataset and 2) a \textbf{H}uman \textbf{C}ollected \textbf{T}actile (\dataabbr) dataset. The SSVTP dataset  (4,587 image-touch pairs) is collected by a UR5 robot, which first captures top-down images from above a work surface on which a set of objects is prearranged, then subsequently presses the DIGIT sensor onto the corresponding location in the workspace. Nonetheless, the SSVTP dataset faces two limitations: 1) its collection in a laboratory environment restricts the diversity of objects, and 2) the asynchronous capture of tactile and visual data can result in misalignments, especially if the object is inadvertently moved by the robot during data acquisition. To address these issues, HCT emphasizes the synchronous acquisition of tactile and visual data to ensure alignment in the captured sensory information.

\dataabbr consists of in-the-wild data visual-tactile data examples collected by 5 humans over 20 total hours using the handheld, 3D-printed data collection device featured in \cref{fig:device}. The device records both visual and tactile observations at 30~Hz. Data frames are collected in ``trajectories" of touches: each trajectory consists of the human approaching, contacting, sliding, and withdrawing from an object with the tactile sensor. We categorize the touch-vision pairs as either in- or out-of-contact with the surface. The visual data are collected at an oblique angle such that the tactile sensor and point of contact are always within the field of view of the camera to preserve vision-touch synchronicity. To improve variety within this dataset, human collectors were instructed to search for interesting and novel real-world tactile examples, such as textures and edges. A small held-out test set (1\% of pairs) from the \dataabbr is hand-annotated, while the rest are pseudo-labeled by GPT-4V, as described in \cref{ssec:language_labeling}.

\subsection{Cleaning Candidate Tactile Images}
\label{ssec:dataset_bg_detect}

We categorize the collected data into in-contact and out-of-contact frames using the pretrained tactile encoder from SSVTP~\cite{kerr2023selfsupervised}. For every touch trajectory, under the assumption that the initial and final frames are out-of-contact, we compute an average of these frames to create a reference background image. This image is then embedded by the pretrained tactile encoder to obtain a latent representation. To determine whether a frame in a touch trajectory is in-contact, we calculate the cosine similarity between its tactile latent embedding and that of the estimated background frame. We consider a tactile frame to be in contact when the cosine similarity falls below 0.6~\cite{kerr2023selfsupervised}. The collected data contains 43,741 pairs of in-contact frames and 169,292 pairs of out-of-contact frames.

\subsection{Language Labeling}
\label{ssec:language_labeling}

\figMethod{t!}
\textbf{Human Labeling}
Since the SSVTP dataset demonstrates strong visual-tactile alignment, we use it as the basis for aligning touch and language as well; we manually annotate the dataset with natural language descriptions of the tactile sensations captured by each data point. We provide human annotators with a tactile vocabulary list of 400 words~\cite{ajbarnett_2023} from which to generate language descriptions of the material properties and tactile feelings of pairs in the SSVTP dataset. These annotators are instructed to choose up to five applicable adjectives that most accurately describe the tactile patterns displayed in each visual-tactile pair.

\textbf{Pseudo-Label Generation with GPT-4V}
We perform pseudo-labeling on the portion of the \dataabbr dataset that is in contact, using GPT-4V to generate language labels describing tactile feelings. We empirically find that providing both the full image and a localized version that is cropped around the point of contact encourages GPT-4V to generate textual labels that are aligned with those of humans, as the full images may contain numerous distractors and out-of-contact objects (see success and failure cases in \cref{fig:dataset}). The specific prompt provided to GPT-4V for pseudo-label generation is reported in \cref{sec:appendix:pseudo}. 

Occasionally, GPT-4V fails or refuses to generate tactile labels for motion blurred or low lighting images. In such cases, we first attempt to generate labels for other images in the same trajectory, then populate the missing labels by randomly sampling from the set of words applied to other in-contact images within the same trajectory. If \textit{no} image in the trajectory can successfully be labeled, that trajectory is excluded from the training portion of the dataset. After this process, we are left with 39,154 pseudo-labeled images.

\subsection{Dataset Statistics}
\label{ssec:dataset_stat}
The SSVTP component contains 4,587 independent image-touch pairs. The \dataabbr component consists of 39,154 newly-collected corresponding in-contact image-tactile frame pairs and 169,292 out-of-contact data pairs. The former dataset contains a unique touch trajectory for each data point, while the latter are collected as 1,486 unique continuous trajectories, each of which consists of one or more contact events with an object of interest. Across both the human- and GPT-4V-labeled portions of the dataset, annotators use 254 unique tactile adjectives. We perform a 99\%-1\% train-test split across both dataset components, with human annotators manually labeling the test set (402 image-touch pairs) for both datasets. On average, GPT-4V uses 4.25 adjectives to describe the tactile sensation on \dataabbr, while human annotators average 2.70 adjectives. A more detailed breakdown of the descriptions is shown in \cref{sec:appendix:distr}.

\section{Tactile-Vision-Language Model}
\label{sec:method}
We first revisit the formulation of ImageBind and ImageBind-LLM. We then describe our pairwise contrastive approach for tactile encoder training, and finally discuss the training recipe of our aligned TVL Model.

\figResults{ht!}

\subsection{Preliminary}
\label{ssec:prelim}
\textbf{ImageBind}~\cite{girdhar2023imagebind} is a multimodal model that learns a joint embedding across six different modalities: images, text, audio, depth, thermal, and IMU data. It utilizes data pairs consisting of vision and one of the other modalities, so that all are ``bound'' to vision. The vision and language encoders are initialized from OpenCLIP~\cite{ilharco_gabriel_2021_5143773} and remain frozen, while the encoders for the other modalities are randomly initialized. Each encoder uses a small, trainable adapter network at the end to project inputs onto a latent space of the same dimension. Encoders are jointly trained through contrastive learning on the normalized latent embeddings using the InfoNCE loss.

\textbf{LLaMA-Adapter}~\cite{zhang2023llamaadapter} and \textbf{ImageBind-LLM}~\cite{han2023imagebindllm} provide efficient instruction finetuning approaches for VLMs, leveraging pretrained multimodal models to encode new modalities. The efficiency of these methods comes from (1) averaging multimodal observations in a single token and (2) a zero-initialized gate that adaptively fuses the multimodal token with the language model. LLaMA-Adapter first pretrains the zero-initialized gate and the projector from the encoder to the language model, then finetunes the language model with LoRA~\cite{hu2022lora}.

\subsection{Tactile Encoder} %
\label{ssec:tac_enc}
In contrast to ImageBind, which independently binds all modalities to vision, we bind each pair of modalities to provide strong supervision for the tactile modality. We calculate contrastive loss between vision-language, tactile-language, and tactile-vision pairs for each data batch. We randomly initialize the tactile encoder as a Vision Transformer (ViT)~\cite{Dosovitskiy2020} and test on three model sizes: ViT-Tiny (5.7M paraeters), ViT-Small (22M), and ViT-Base (86M). We notice that directly adopting the ImageBind training recipe leads to overfitting the relatively small training dataset of 44K pairs of in-contact data. Contrary to prior works~\cite{kerr2023selfsupervised, yang2022touch, dave2024multimodal}, we find that leveraging data in which the tactile sensor is not in contact with a surface (background images) can mitigate this overfitting problem and enhance tactile representation learning by improving visual data diversity (see~\cref{fig:appendix_contact} in appendix). Therefore, we ensure that for a fraction $\gamma=10\%$ of the training data, the sensor is not in contact, and we assign these examples a text label of ``background''. In addition, we remove the projectors from the vision and language encoders, so that the tactile encoder directly projects to the common latent space of the original CLIP. Finally, to increase the diversity of language labels, we randomly shuffle and select a subset of the words in the tactile description for each image. Together, these methods help to mitigate overfitting (refer to \cref{sec:appendix:overfit}).

\subsection{Alignment with Language Models}
\label{ssec:language_align}
We follow the two-stage training proposed in ImageBind-LLM~\cite{han2023imagebindllm}, exchanging the ImageBind encoders with \algabbr encoders. We pretrain on both the LLaVA Visual Instruct CC3M~\cite{liu2023llava} 595K subset and the \algabbr dataset. For the CC3M subset, we provide an empty tactile image to the tactile modality. During finetuning, we use a combination of \algabbr, Alpaca~\cite{alpaca} and LLaVA Visual Instruct 150K~\cite{liu2023llava}. Empirically, we find that training our dataset alone is not sufficient to overcome the safety fine-tuning of LLaMA2~\cite{touvron2023llama}, resulting in the model's refusal to answer questions regarding tactile sensations. Details on the prompts for \algabbr for instruction fine-tuning is in \cref{sec:appendix:vqa_prompts}.

\section{Experiments}
\label{sec:experiments}
We quantitatively assess the multimodal capabilities of the \algabbr model in two experimental settings: a cross-modal classification task and a tactile-semantic description task.
\tabEncExp{t}
\begin{table*}
\centering
{\begin{tabular}{r|ccccc|cccccc}
\toprule[1pt]
& & \multicolumn{3}{c}{\textbf{Encoder Pre-training Modalities}} & & & \multicolumn{3}{c}{\textbf{Score} (1-10)} & $p$-value\\
\cline{3-5}\cline{8-10}
& & Vision & Tactile & Language & & & SSVTP & \dataabbr & \algabbr & $(\textrm{d.f.}=401)$\\
\midrule[0.1pt]
LLaVA-1.5 7B & & \checkmark & - & \checkmark & & & 3.64 & 3.55 & 3.56 & $1.21 \times 10^{-9}$ \\
LLaVA-1.5 13B & & \checkmark & - & \checkmark & & & 3.55 & 3.63 & 3.62 & $1.49 \times 10^{-9}$\\
ViP-LLaVA 7B & & \checkmark & - & \checkmark & & & 2.72 & 3.44 & 3.36 & $8.77 \times 10^{-16}$\\
ViP-LLaVA 13B & & \checkmark & - & \checkmark & & & 4.10 & 3.76 & 3.80 & $1.72 \times 10^{-6}$\\
LLaMA-Adapter & & \checkmark & - & \checkmark & & & 2.56 & 3.08 & 3.02 & $2.68 \times 10^{-17}$ \\
BLIP-2 Opt-6.7b & & \checkmark & - & \checkmark & & & 2.02 & 2.72 & 2.64 & $1.92 \times 10^{-31}$\\
InstructBLIP 7B & & \checkmark & - & \checkmark & & & 1.40 & 1.30 & 1.31 & $1.07 \times 10^{-84}$\\
InstructBLIP 13B & & \checkmark & - & \checkmark & & & 1.44 & 1.21 & 1.24 & $4.64 \times 10^{-88}$\\
GPT-4V & & \checkmark & - & \checkmark & & & 5.02 & 4.42 & 4.49 & - \\
\midrule[0.1pt]
SSVTP-LLaMA & & \checkmark & \checkmark & - & & & 2.58 & 3.67 & 3.54 & $1.79 \times 10^{-9}$ \\
\midrule[0.1pt]
\algabbr-LLaMA (ViT-Tiny) & & \checkmark & \checkmark & \checkmark & & & 6.09 & 4.79 & 4.94 & $4.24 \times 10^{-5}$\\
\algabbr-LLaMA (ViT-Small) & & \checkmark & \checkmark & \checkmark & & & 5.81 & 4.77 & 4.89 & $6.02 \times 10^{-4}$\\
\algabbr-LLaMA (ViT-Base) & & \checkmark & \checkmark & \checkmark & & & \textbf{6.16} & \textbf{4.89} & \textbf{5.03} & $3.46 \times 10^{-6}$\\
\bottomrule[1pt]
\end{tabular}}
\caption{\textbf{TVL Benchmark Performance.} We benchmarked \algabbr-LLaMA against existing VLMs and SSVTP-LLaMA, a model fine-tuned using SSVTP tactile-vision encoders, for generating tactile descriptions from tactile-image observations, and used GPT-4 to numerically score the performance on each constituent part of the \algabbr test set. We report $p$-values from two-sided paired sample $t$-tests on each model's scores against GPT-4V's scores on the tactile-semantic task.}
\label{tab:evaluation}
\end{table*}

\subsection{Evaluation \& Metrics}
\label{ssec:eval_metrics}
\textbf{Open Vocabulary Tactile Classification}
We cast the human-labeled \algabbr test set as a 402-way classification problem and evaluate the tactile encoder's performance by measuring the top-1 and top-5 accuracy for both tactile-vision and tactile-language classification. 
Since many tactile observations can be described in multiple semantically similar ways (\emph{e.g.} rigid is synonymous with stiff) and CLIP language embedding is not permutation invariant (\emph{e.g.} ``soft, smooth" and ``smooth, soft" have different embeddings), we propose an alternative method to calculate the ground truth labels for tactile-language classification. %

We first prompt GPT-4 to generate a set of 5 (the average length of tactile pseudo-labels) synonyms for each word in the set of descriptors used by the human annotators of the SSVTP dataset, resulting in 799 distinct adjectives describing tactile sensations. We obtain the CLIP language embedding for these adjectives and calculate the cosine similarities of each original descriptor with each of its generated synonyms. We consider the minimum $\phi$ of these cosine similarities to be a threshold for semantically similar vocabulary. 
For each tactile image, we define the set of correct language labels as all labels in the test set whose cosine similarity with the image's original language label exceeds $\phi$. 
Using these labels, we calculate the top-1 and top-5 accuracy. Empirically, we find $\phi = 0.636$. We also report top-1 and top-5 accuracy using the 25th, 50th, and 75th percentile of the cosine similarities as the threshold in \cref{tbl:enc_tac_text}.

\textbf{TVL Benchmark}
We evaluate the capabilities of LLMs to generate tactile descriptions on the \algabbr test set. Given a visual input image, a cropped visual image centered on the tactile sensor, and a corresponding tactile image, we ask the model to describe the tactile sensations of the object in question with a set of no more than 5 adjectives.

To obtain a numerical comparison, we prompt text-only GPT-4 to score the similarity of the model's response against human-annotated ground truth semantic labels on a scale of 1 to 10 (where a higher score indicates better instruction-following and a closer descriptive match), as well as to explain the score given, similar to prior works~\cite{liu2023llava, vicuna2023}. A sample of model outputs is provided in \cref{fig:results}, and prompts used for generation and evaluation are reported in \cref{sec:appendix:pseudo}. We compare against existing open-source VLMs~\cite{liu2023improvedllava, cai2023vipllava, li2023blip2, dai2023instructblip} and GPT-4V. As an additional baseline, we use the SSVTP~\cite{kerr2023selfsupervised} tactile and image encoder to finetune the language model; we call the resulting model SSVTP-LLaMA.

\subsection{Results}
\begin{table*}[t]
\newlength\savewidth\newcommand\shline{\noalign{\global\savewidth\arrayrulewidth
  \global\arrayrulewidth 1pt}\hline\noalign{\global\arrayrulewidth\savewidth}}
\newcommand{\tablestyle}[2]{\setlength{\tabcolsep}{#1}\renewcommand{\arraystretch}{#2}\centering\ftsize}
\renewcommand{\paragraph}[1]{\vspace{1.25mm}\noindent\textbf{#1}}
\definecolor{baselinecolor}{gray}{.9}
\newcommand{\baseline}[1]{\cellcolor{baselinecolor}{#1}}

\newcolumntype{x}[1]{>{\centering\arraybackslash}p{#1pt}}
\newcolumntype{y}[1]{>{\raggedright\arraybackslash}p{#1pt}}
\newcolumntype{z}[1]{>{\raggedleft\arraybackslash}p{#1pt}}

\vspace{-.2em}
\centering
\ftsize
\subfloat[
\ftsize
\textbf{Model Architecture} used for transformer encoder backbone. 
\label{tab:ablation-architecture}
]{
\centering
\ftsize
\begin{minipage}{0.30\linewidth}{\begin{center}
\begin{tabular}{y{45}x{30}x{30}}
& Tac./Text & Tac./Vis. \\
Model & \% Acc. & \% Acc. \\
\shline
ViT-Tiny & \textbf{36.7} & 79.5 \\
ViT-Small & \baseline{36.3} & \baseline{78.0} \\
ViT-Base & 30.7 & \textbf{81.7} \\
\end{tabular}
\end{center}}\end{minipage}
}
\hspace{1em}
\subfloat[
\ftsize
\textbf{Disable Tactile-Text Loss.} ImageBind-style training, lacking direct supervision for tactile and language alignment, reduces model accuracy.
\label{tab:ablation-imagebind}
]{
\centering
\ftsize
\begin{minipage}{0.30\linewidth}{\begin{center}
\begin{tabular}{y{35}x{30}x{30}}
Tactile- & Tac./Text & Tac./Vis. \\
Text Loss & \% Acc. & \% Acc. \\
\shline
Enabled & \baseline{\textbf{36.3}} & \baseline{78.0} \\
Disabled & 20.3 & \textbf{81.6} \\
\\
\end{tabular}
\end{center}}\end{minipage}
}
\hspace{1em}
\subfloat[
\ftsize
\textbf{Modality-Specific Training.} Contrastive losses across all modalities improve performance.
\label{tab:ablation-modality}
]{
\begin{minipage}{0.30\linewidth}{\begin{center}
\begin{tabular}{y{30}x{30}x{30}}
& Tac./Text & Tac./Vis. \\
Modality & \% Acc. & \% Acc. \\
\shline
All & \baseline{\textbf{36.3}} & \baseline{78.0} \\
\textminus Vision & 29.9 & 1.0 \\
\textminus Text & 21.5 & \textbf{85.8}
\end{tabular}
\end{center}}\end{minipage}
}

\centering
\vspace{.7em}
\subfloat[
\ftsize
\textbf{Contact Data Mix.} Adding non-contact frames to the training data does not significantly improve performance.
\label{tab:ablation-contact}
]{
\begin{minipage}[t]{0.30\linewidth}{\begin{center}
\begin{tabular}{y{45}x{30}x{30}}
& Tac./Text & Tac./Vis. \\
Contact & \% Acc. & \% Acc. \\
\shline
Contact & 36.2 & \textbf{80.1} \\
+ 10\% N.C. & \baseline{\textbf{36.3}} & \baseline{78.0} 
\\
\\
\end{tabular}
\end{center}}\end{minipage}
}
\hspace{1em}
\subfloat[
\ftsize
\textbf{Prompting.} TVL Performance does not depend strongly on prompt formatting.
\label{tab:ablation-prompting}
]{
\begin{minipage}[t]{0.30\linewidth}{\begin{center}
\begin{tabular}{y{35}x{30}x{30}}
& Tac./Text & Tac./Vis. \\
Prompting & \% Acc. & \% Acc. \\
\shline
Baseline & \baseline{36.3} & \baseline{78.0} \\
+ Prompt & \textbf{37.7} & \textbf{78.7} \\
\\
\end{tabular}
\end{center}}\end{minipage}
}
\hspace{1em}
\subfloat[
\ftsize
\textbf{Training Dataset.} Models which are exposed to the \dataabbr dataset in training outperform SSVTP-only models.
\label{tab:ablation-dataset}
]{
\begin{minipage}[t]{0.30\linewidth}{
\begin{center}
\begin{tabular}{y{30}x{30}x{30}}
& Tac./Text & Tac./Vis. \\
Dataset & \% Acc. & \% Acc. \\
\shline
SSVTP & 19.2 & 8.0 \\
\dataabbr & \textbf{38.4} & 74.4 \\
\algabbr & \baseline{36.3} & \baseline{\textbf{78.0}}
\end{tabular}
\end{center}}\end{minipage}
}
\setlength{\abovecaptionskip}{10pt plus 3pt minus 2pt}
\caption{\ftsize \textbf{Ablations and Sensitivity Analysis} for the \algabbr tactile encoder. We report top-1 and top-5 tactile-text and tactile-vision classification accuracy with ViT-Small. \colorbox{baselinecolor}{baseline} indicates the default setting for training the \algabbr tactile encoder, which is the best-performing model on the \emph{validation set} unless noted otherwise. \textbf{Bold} indicates the highest accuracy on the \emph{test set}. 
Such discrepancy in performance is described in \cref{ssec:ablation}.
}
\label{tab:ablation}
\vspace{-15pt}
\end{table*}

\textbf{Classification}
\label{sssection:classification}
We summarize the tactile classification task results in \cref{tab:encoder}. Because we use OpenCLIP to encode image and language observations, the \algabbr encoder shares its vision-language accuracy scores with OpenCLIP. We compare the tactile-vision accuracy of our encoder against \citet{kerr2023selfsupervised}; because they train on a small dataset collected in a lab setup, their model performs well on the SSVTP dataset, but does not generalize well to the new ``in-the-wild" dataset. Since the tactile encoder is aligned to the language description of tactility, it shows better tactile-text alignment than OpenCLIP's vision-text alignment.

\textbf{TVL Benchmark}
\label{sssection:generation}
We present summary statistics for the tactile-semantic generation results in \cref{tab:evaluation}. We find that open-source VLMs perform worse than GPT-4V on the proposed benchmark, likely due to the limited diversity and lack of focus on human tactility in the visual data that they have been trained on. On the other hand, all versions of \algabbr-LLaMA outperform GPT-4V, suggesting that the trained models can generalize beyond the small fraction of human labels provided as part of the dataset. Both these findings are statistically significant at the $\alpha=0.05$ level.
Results also suggest that tactile-language alignment is necessary, as evidenced by the lower score of SSVTP-LLaMA, which only uses tactile and vision modalities during pre-training. 

Overall, our experiments suggest that: 1) the \algabbr tactile encoder trained on the \algabbr dataset is aligned with the language latent space and scores higher (+29\%) on the classification task as compared to visual-tactile pretrained encoders and generic vision-language encoders (OpenCLIP); and 2) \algabbr-LLaMA models trained to generate tactile language descriptions from visual and tactile observations more closely match human descriptions on the novel TVL Benchmark (at least +12\%) compared to existing VLMs.

\subsection{Ablations}
\label{ssec:ablation}
This section presents six ablation and sensitivity analyses shown in \cref{tab:ablation} examining the impact of model size and the proposed dataset on the encoder's multi-modal classification performance.
More ablations are included in the appendix.

\textbf{Model Sizes} (\cref{tab:ablation-architecture}) Performance varies significantly among different encoder sizes. ViT-Base has the highest validation accuracy but lags on the test set due to distribution shifts: the training labels from GPT-4V are less detailed and accurate compared to human-annotated test data. However, in tactile-vision classification on synchronized data, ViT-Base outperforms both of the smaller models.

\textbf{Disable Tactile-Text Loss} (\cref{tab:ablation-imagebind}) resembles the setup in ImageBind~\cite{girdhar2023imagebind}, where data in all three modalities are considered but the tactile-text loss is omitted. Results suggest that using language to supervise the tactile encoder better aligns those two modalities. 

\textbf{Data} (Tables \hyperref[tab:ablation-modality]{3c-f}) We perform four sensitivity analyses on the different compositions of the dataset for training. We find that leveraging data from all three modalities improves tactile-language alignment. While adding not-in-contact data prevents the model from overfitting to the training set, its test set performance is comparable with having only in-contact data. We also experimented with prompting used in vanilla CLIP training~\cite{radford2021learning}, which brings marginal improvements in accuracy. Lastly, we separately train the model on SSVTP and \dataabbr, and we find that the pseudo-labeled dataset can provide comparable performance with training on the entire dataset, which suggests that \algabbr's tactile encoder can effectively leverage self-supervised learning to reduce the dependency on large, fully-labeled datasets while maintaining task performance.

\section{Discussion and Conclusion}

The research presented has several limitations. 
While the study highlights the use of VLMs for labeling tactile data, the distinct nature of touch compared to visual perception suggests a limit to the accuracy of tactile labels derived solely from vision. 
Due to the data collection hardware, the camera may not have an unoccluded view of the surface or object that the tactile sensor contacts, which may increase the difficulty of aligning touch with vision and reduce the quality of pseudo-labels generated from images. We hope that future research can further increase the scale of touch-vision-language datasets to improve multimodal alignment.

In sum, to align the tactile and language modalities, this work introduces \algabbr, a dataset that features tactile, vision, and tactile-semantic descriptions. Utilizing the dataset, we train a tactile encoder that is aligned to both vision and natural language. We demonstrate that by using the trained tactile encoder, \algabbr-LLaMA can generate tactile descriptions in natural language that align more closely with human descriptions than those generated by existing VLMs.

\section{Impact Statements}
The data present in this paper is anonymized. This work could benefit future large generative models also considering touch as a sensing modality and can be useful for researchers studying pseudo-label-based learning methods. At the same time, the model introduced will contribute to achieving a better digitalization of touch and the use of touch in robotics. This paper presents work whose goal is to advance the field of Machine Learning. There are many potential societal benefits of our work, 
none of which we feel must be specifically highlighted here. 

\section{Acknowledgments}
This research was supported as a BAIR Open Research Common Project with Meta. This research was performed at the AUTOLAB at UC Berkeley in affiliation with the Berkeley AI Research (BAIR) Lab, and the CITRIS "People and Robots" (CPAR) Initiative.
In their academic roles at UC Berkeley, Letian Fu, Gaurav Datta, Huang Huang, William Chung-Ho Panitch, Jaimyn Drake, and Ken Goldberg are supported in part by donations from Meta, Google, Autodesk, Siemens, Toyota Research Institute, Bosch, and by equipment grants from PhotoNeo, Nvidia, and Intuitive Surgical.
Roberto Calandra is funded by the German Research Foundation (DFG, Deutsche Forschungsgemeinschaft) as part of Germany’s Excellence Strategy – EXC 2050/1 – Project ID 390696704 – Cluster of Excellence “Centre for Tactile Internet with Human-in-the-Loop” (CeTI) of Technische Universität Dresden, and by Bundesministerium für Bildung und Forschung (BMBF) and German Academic Exchange Service (DAAD) in project 57616814 (\href{https://secai.org/}{SECAI}, \href{https://secai.org/}{School of Embedded and Composite AI}). We thank Justin Kerr, Chung Min Kim, Ryan Hoque, and Xudong Wang for their helpful discussions and feedback.

\bibliography{main}
\bibliographystyle{icml2024}

\clearpage
\appendix
\def\tabTacEncHyper#1{
    \begin{table}[#1]
    \centering
    \begin{tabular}{cc}
    \toprule
    Config                 & Value                                                                             \\ \hline
    optimizer              & AdamW~\cite{loshchilov2017decoupled}                                                                             \\
    base learning rate     & 1.5e-4                                                                            \\
    learning rate schedule & cosine decay~\cite{loshchilov2016sgdr}                                                                      \\
    batch size             & 4096                                                                              \\
    weight decay           & 0.05                                                                              \\
    optimizer momentum     & $\beta_1, \beta_2$ = 0.9, 0.95~\cite{chen2020generative}                                                      \\
    warm up epoch~\cite{Goyal2017b}          & 20, 40                                                                            \\
    total epochs           & 400, 800                                                                          \\
    augmentation           & \begin{tabular}[c]{@{}c@{}}RandomResizedCrop,\\ RandomHorizontalFlip\end{tabular} \\
    \bottomrule
    \end{tabular}
    \caption{Pretraining Hyperparameters}
    \label{tbl:pretrain_hyper}
    \end{table}
}
\def\tabBGS#1{
    \begin{table}[#1]
    \centering
    \begin{tabular}{rccc}
    \toprule 
     & \begin{tabular}[c]{@{}c@{}}Tac./Text\\ \% Acc\end{tabular} & \begin{tabular}[c]{@{}c@{}}Tac./Vis\\ \% Acc\end{tabular}  & \begin{tabular}[c]{@{}c@{}}TVL\\ Score\end{tabular}  \\ \hline
    In-Contact Frames  & 36.2                                                       & \textbf{80.1}                                     & 4.81                 \\
    +10\% No-Contact     & 36.3                                                       & 78.0                                              & 4.89        \\
    + Background Subtract & \textbf{42.3}                                                       & 78.9                                     & \textbf{5.06}                \\
    \bottomrule
    \end{tabular}
    \caption{Effect of no-contact data and background subtraction during ViT-Small tactile encoder training on classification accuracy and performance on the TVL benchmark.}
    \label{tab:background}
    \end{table}
}

\def\tabZeroShot#1{
    \begin{table}[#1]
    \centering
    \ssmall
    \begin{tabular}{lccc}
    \hline
                                                                                                & \begin{tabular}[c]{@{}c@{}}Zero-Shot \\ Tactile\end{tabular} & \begin{tabular}[c]{@{}c@{}}Zero-Shot \\ Vision\end{tabular} & \begin{tabular}[c]{@{}c@{}}Tactile \\ \& Vision\end{tabular} \\ \hline
    \begin{tabular}[c]{@{}l@{}}TVL-LLaMA \\ (ViT-Tiny)\end{tabular}                             & \textbf{4.56}                                                         & 4.66                                                        & 4.94                                                         \\
    \begin{tabular}[c]{@{}l@{}}TVL-LLaMA \\ (ViT-Small)\end{tabular}                            & 3.50                                                         & 4.81                                                        & 4.89                                                         \\
    \begin{tabular}[c]{@{}l@{}}TVL-LLaMA \\ (ViT-Base)\end{tabular}                             & 2.80                                                         & \textbf{4.85}                                                        & 5.03                                                         \\ \hline
    \begin{tabular}[c]{@{}l@{}}TVL-LLaMA \\ (ViT-Small)\\ + Background Subtract\end{tabular} & 4.52                                                         & -                                                            & \textbf{5.06}                                                         \\ \hline
    \end{tabular}
    \caption{Dropping one modality (out-of-distribution) zero shot experiments}
    \label{tab:ood_zeroshot}
    \end{table}
}

\def\tabPerDatasetSSVTP#1{
    \begin{table}[#1]
\centering
{\begin{tabular}{r|cc}
\toprule[1pt]
& \textbf{Score} & $p$-value \\
& (1-10) & $(\mathrm{d.f.}=401)$ \\
\midrule[0.1pt]
LLaVA-1.5 7B & 3.64 & $2.32 \times 10^{-3}$ \\
LLaVA-1.5 13B & 3.55 & $1.30\times 10^{-3}$\\
ViP-LLaVA 7B & 2.72  & $4.45 \times 10^{-8}$\\
ViP-LLaVA 13B & 4.10 & $3.76 \times 10^{-2}$\\
LLaMA-Adapter & 2.56 & $7.826 \times 10^{-6}$ \\
BLIP-2 Opt-6.7b & 2.02 & $2.74 \times 10^{-9}$\\
InstructBLIP 7B & 1.40 & $1.49 \times 10^{-13}$\\
InstructBLIP 13B & 1.44 & $4.68 \times 10^{-14}$\\
GPT-4V & 5.02 & - \\
\midrule[0.1pt]
SSVTP-LLaMA & 2.58 & $9.33 \times 10^{-6}$ \\
\midrule[0.1pt]
\algabbr-LLaMA (ViT-Tiny) & 6.09 & $2.65 \times 10^{-2}$\\
\algabbr-LLaMA (ViT-Small) & 5.81 & $1.02 \times 10^{-1}$\\
\algabbr-LLaMA (ViT-Base) & \textbf{6.16} & $1.67 \times 10^{-2}$\\
\bottomrule[1pt]
\end{tabular}}
\caption{\textbf{TVL Benchmark Performance on SSVTP.} We benchmarked \algabbr-LLaMA against existing VLMs and SSVTP-LLaMA, and show here the performance on only the SSVTP dataset. We report $p$-values from two-sided paired sample $t$-tests on each model's scores against GPT-4V's scores.}
\label{tab:appendix_ssvtp_gen}
\end{table}
}

\def\tabPerDatasetHCT#1{
    \begin{table}[#1]
\centering
{\begin{tabular}{r|cc}
\toprule[1pt]
& \textbf{Score} & $p$-value \\
& (1-10) & $(\mathrm{d.f.}=401)$ \\
\midrule[0.1pt]
LLaVA-1.5 7B & 3.55 & $8.49 \times 10^{-8}$ \\
LLaVA-1.5 13B & 3.63 & $1.74 \times 10^{-7}$\\
ViP-LLaVA 7B & 3.44 & $4.10 \times 10^{-11}$\\
ViP-LLaVA 13B & 3.76 & $1.57 \times 10^{-5}$\\
LLaMA-Adapter & 3.08 & $2.05 \times 10^{-13}$\\
BLIP-2 Opt-6.7b & 2.72 & $1.25 \times 10^{-24}$\\
InstructBLIP 7B &  1.30 & $8.02 \times 10^{-73}$\\
InstructBLIP 13B & 1.21 & $9.74 \times 10^{-76}$\\
GPT-4V & 4.42 & - \\
\midrule[0.1pt]
SSVTP-LLaMA & 3.67 & $3.24 \times 10^{-6}$ \\
\midrule[0.1pt]
\algabbr-LLaMA (ViT-Tiny) & 4.79 & $5.79 \times 10^{-4}$\\
\algabbr-LLaMA (ViT-Small) & 4.77 & $2.64 \times 10^{-3}$\\
\algabbr-LLaMA (ViT-Base) & \textbf{4.89} & $6.82 \times 10^{-5}$\\
\bottomrule[1pt]
\end{tabular}}
\caption{\textbf{TVL Benchmark Performance on \dataabbr.} We benchmarked \algabbr-LLaMA against existing VLMs and SSVTP-LLaMA, and show here the performance on only the \dataabbr dataset. We report $p$-values from two-sided paired sample $t$-tests on each model's scores against GPT-4V's scores.}
\label{tab:appendix_hct_gen}
\end{table}
}

\def\tabtactilestat#1{
    \begin{table}[#1]
    \centering
    \begin{tabular}{lcc}
    \hline
    Tactile Statistics    & Mean                                                             & Std.                                                           \\ \hline
    With Background       & \begin{tabular}[c]{@{}c@{}}0.292\\ 0.297\\ 0.291\end{tabular}    & \begin{tabular}[c]{@{}c@{}}0.188\\ 0.195 \\ 0.219\end{tabular} \\ \hline
    Background Subtracted & \begin{tabular}[c]{@{}c@{}}-0.008\\ -0.019\\ -0.018\end{tabular} & \begin{tabular}[c]{@{}c@{}}0.045\\ 0.044\\ 0.053\end{tabular}  \\ \hline
    \end{tabular}
    \caption{Tactile Normalization Statistics}
    \label{tab:appendix_tacstat}
\end{table}
}

\def\tabrgbstat#1{
    \begin{table}[#1]
    \centering
    \begin{tabular}{lcc}
    \hline
    Image Statistics    & Mean                                                          & Std.                                                          \\ \hline
    OpenCLIP Statistics & \begin{tabular}[c]{@{}c@{}}0.481\\ 0.458\\ 0.408\end{tabular} & \begin{tabular}[c]{@{}c@{}}0.269\\ 0.261\\ 0.276\end{tabular} \\ \hline
    \end{tabular}
    \caption{RGB Normalization Statistics}
    \label{tab:appendix_rgbstat}
    \end{table}
}

\def\tabenchyper#1{
    \begin{table*}[#1]
    \centering
    \begin{tabular}{cc}
    \toprule
    Config                 & Value                                                                             \\ \hline
    optimizer              & AdamW~\cite{loshchilov2017decoupled}                                                                             \\
    base learning rate     & 1.5e-4                                                                            \\
    learning rate schedule & cosine decay~\cite{loshchilov2016sgdr}                                                                      \\
    batch size             & 256                                                                              \\
    weight decay           & 0.05                                                                              \\
    optimizer momentum     & $\beta_1, \beta_2$ = 0.9, 0.95~\cite{chen2020generative}                                                      \\
    warm up epoch~\cite{Goyal2017b}          & 10                                                                            \\
    total epochs           & 200                                                                          \\
    RGB Augmentation           & \begin{tabular}[c]{@{}c@{}}RandomHorizontalFlip,\\ ColorJitter,\\RandomGrayscale,\\GaussianBlur\end{tabular}\\
    Tactile Augmentation   & (Optional) Background Subtraction \\
    \bottomrule
    \end{tabular}
    \caption{Encoder Pretraining Hyperparameters}
    \label{tbl:enc_hyper}
    \end{table*}
}

\def\tabTVLMPre#1{
    \begin{table*}[#1]
    \centering
    \begin{tabular}{cc}
    \toprule
    Config                 & Value                                                                             \\ \hline
    optimizer              & AdamW~\cite{loshchilov2017decoupled}                                                                             \\
    base learning rate     & 1.5e-4                                                                            \\
    learning rate schedule & cosine decay~\cite{loshchilov2016sgdr}                                                                      \\
    batch size             & 256                                                                              \\
    weight decay           & 0.05                                                                              \\
    optimizer momentum     & $\beta_1, \beta_2$ = 0.9, 0.95~\cite{chen2020generative}                                                      \\
    warm up epoch~\cite{Goyal2017b}          & 10                                                                            \\
    total epochs           & 200                                                                          \\
    RGB Augmentation           & \begin{tabular}[c]{@{}c@{}}RandomHorizontalFlip,\\ ColorJitter,\\RandomGrayscale,\\GaussianBlur\end{tabular}\\
    Tactile Augmentation   & (Optional) Background Subtraction \\
    \bottomrule
    \end{tabular}
    \caption{\algabbr-LLaMA Pretraining Hyperparameters}
    \label{tbl:appendix_tvlm_pre}
    \end{table*}
}

\def\tabenctactext#1{
\begin{table*}[#1]
\centering
\begin{tabular}{cr|lcclccccc}
\hline
\multicolumn{1}{l}{\multirow{2}{*}{Percentile}} & \multicolumn{1}{c|}{\multirow{2}{*}{}} &  & \multicolumn{2}{c}{SSVTP} &  & \multicolumn{2}{c}{\dataabbr} &  & \multicolumn{2}{c}{\algabbr} \\ \cline{4-5} \cline{7-8} \cline{10-11} 
\multicolumn{1}{l}{}                          & \multicolumn{1}{c|}{}                  &  & Top-1       & Top-5       &  & Top-1      & Top-5      &  & Top-1      & Top-5     \\ \hline
\multirow{3}{*}{0}                            & ViT-Tiny                               &  & 29.4\%      & 71.7\%      &  & 34.8\%     & 70.1\%     &  & 36.7\%     & 70.3\%    \\
                                              & ViT-Small                                  &  & 42.4\%      & 76.1\%      &  & 36.5\%     & 68.0\%     &  & 36.3\%     & 66.4\%    \\
                                              & ViT-Base                                  &  & 38.0\%      & 69.6\%      &  & 34.8\%     & 65.6\%     &  & 30.7\%     & 63.6\%    \\  \hline
\multirow{3}{*}{25}                           & ViT-Tiny                               &  & 3.3\%       & 21.7\%      &  & 7.2\%      & 22.9\%     &  & 4.6\%      & 14.1\%    \\
                                              & ViT-Small                                  &  & 10.9\%      & 33.7\%      &  & 9.1\%      & 21.5\%     &  & 6.7\%      & 19.5\%    \\
                                              & ViT-Base                                  &  & 8.7\%       & 31.5\%      &  & 5.9\%      & 14.0\%     &  & 4.4\%      & 13.7\%    \\ \hline
\multirow{3}{*}{50}                           & ViT-Tiny                               &  & 3.3\%       & 19.6\%      &  & 4.8\%      & 17.8\%     &  & 3.7\%      & 11.8\%    \\
                                              & ViT-Small                                  &  & 10.9\%      & 32.6\%      &  & 6.6\%      & 15.3\%     &  & 5.9\%      & 11.0\%    \\
                                              & ViT-Base                                  &  & 7.6\%       & 28.3\%      &  & 4.5\%      & 9.8\%      &  & 3.5\%      & 11.0\%    \\ \hline
\multirow{3}{*}{75}                           & ViT-Tiny                               &  & 3.3\%       & 19.6\%      &  & 4.1\%      & 14.2\%     &  & 3.7\%      & 10.7\%  \\
                                              & ViT-Small                                  &  & 10.9\%      & 28.3\%      &  & 3.5\%      & 7.9\%      &  & 3.4\%      & 10.2\%    \\
                                              & ViT-Base                                  &  & 7.6\%       & 28.3\%      &  & 3.5\%      & 7.9\%      &  & 3.4\%      & 10.2\%    \\ \hline
\end{tabular}
\caption{Effect of Model Architecture and Similarity Threshold $\phi$ on \textbf{Tactile-Text} Classification Accuracy. The similarity thresholds $\phi$ for each percentile are 0.636 (0th), 0.859 (25th), 0.893 (50th), and 0.921 (75th).}
\label{tbl:enc_tac_text}
\end{table*}
}

\def\tabenctacvis#1{
\begin{table*}[#1]
\centering
\begin{tabular}{r|ccccccccc}
\hline
         &  & \multicolumn{2}{c}{SSVTP} &  & \multicolumn{2}{c}{\dataabbr} &  & \multicolumn{2}{c}{\algabbr} \\ \cline{3-4} \cline{6-7} \cline{9-10} 
         &  & Top-1        & Top-5       &  & Top-1      & Top-5     &  & Top-1          & Top-5          \\ \hline
ViT-Tiny &  & 34.8\%       & 70.7\%      &  & 85.3\%     & 99.0\%    &  & 79.5\%         & 95.7\%         \\
ViT-Small    &  & 28.3\%       & 69.6\%      &  & 84.4\%     & 98.9\%    &  & 78.0\%         & 95.2\%         \\
ViT-Base    &  & 34.8\%       & 66.3\%      &  & 87.8\%     & 99.7\%    &  & 81.7\%         & 95.7\%         \\ \hline
\end{tabular}
\caption{Effect of Tactile Encoder Model Architecture on \textbf{Tactile-Vision} Classification. }
\label{tbl:enc_tacvis}
\end{table*}
}

\def\figAppendixScaling#1{
    \begin{figure}[#1]
    \centering
    \begin{subfigure}[b]{0.45\textwidth}
        \centering
        \includegraphics[width=\textwidth]{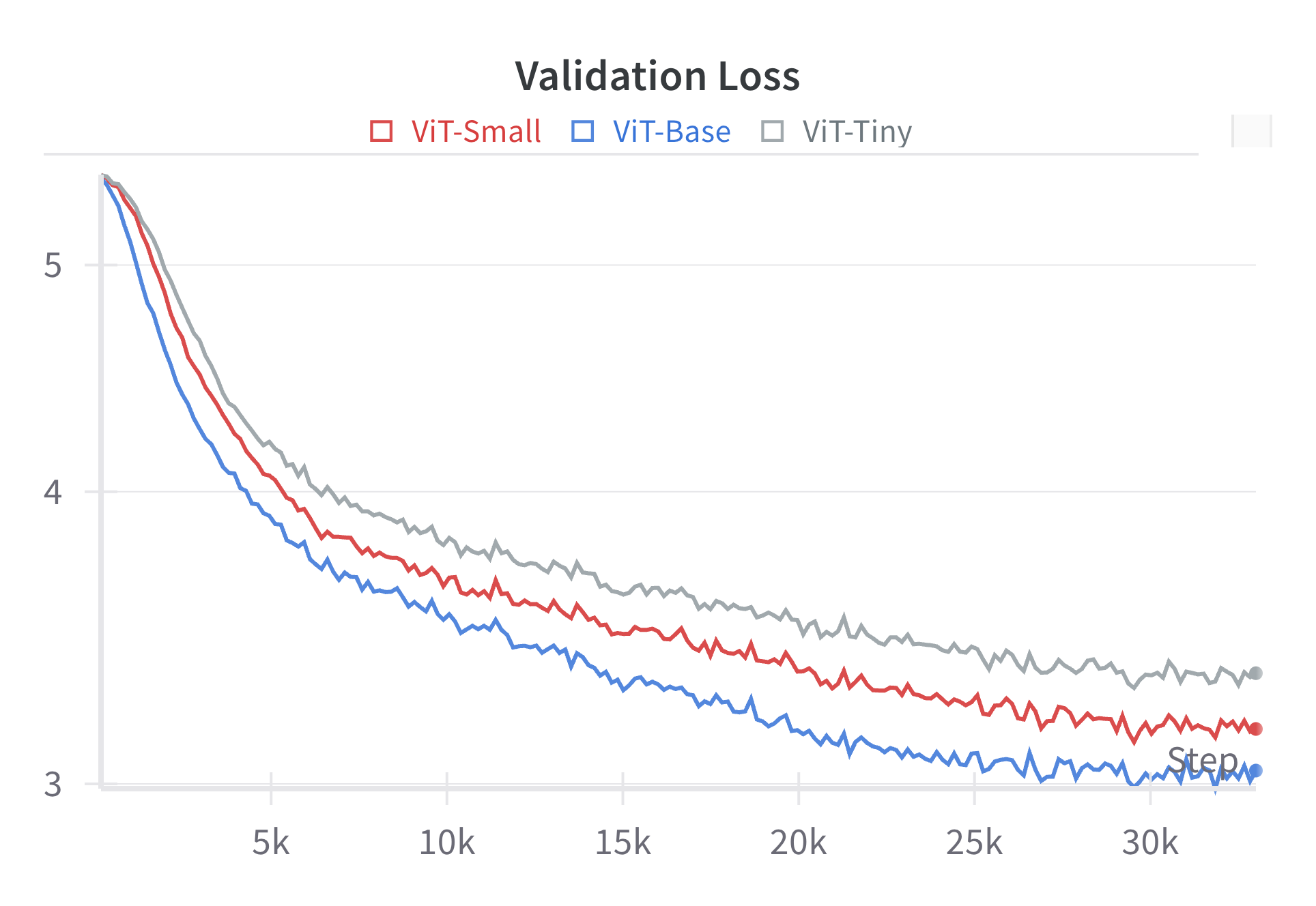}
        \label{fig:vit_vs_valloss}
    \end{subfigure}
    \hfill 
    \begin{subfigure}[b]{0.45\textwidth}
        \centering
        \includegraphics[width=\textwidth]{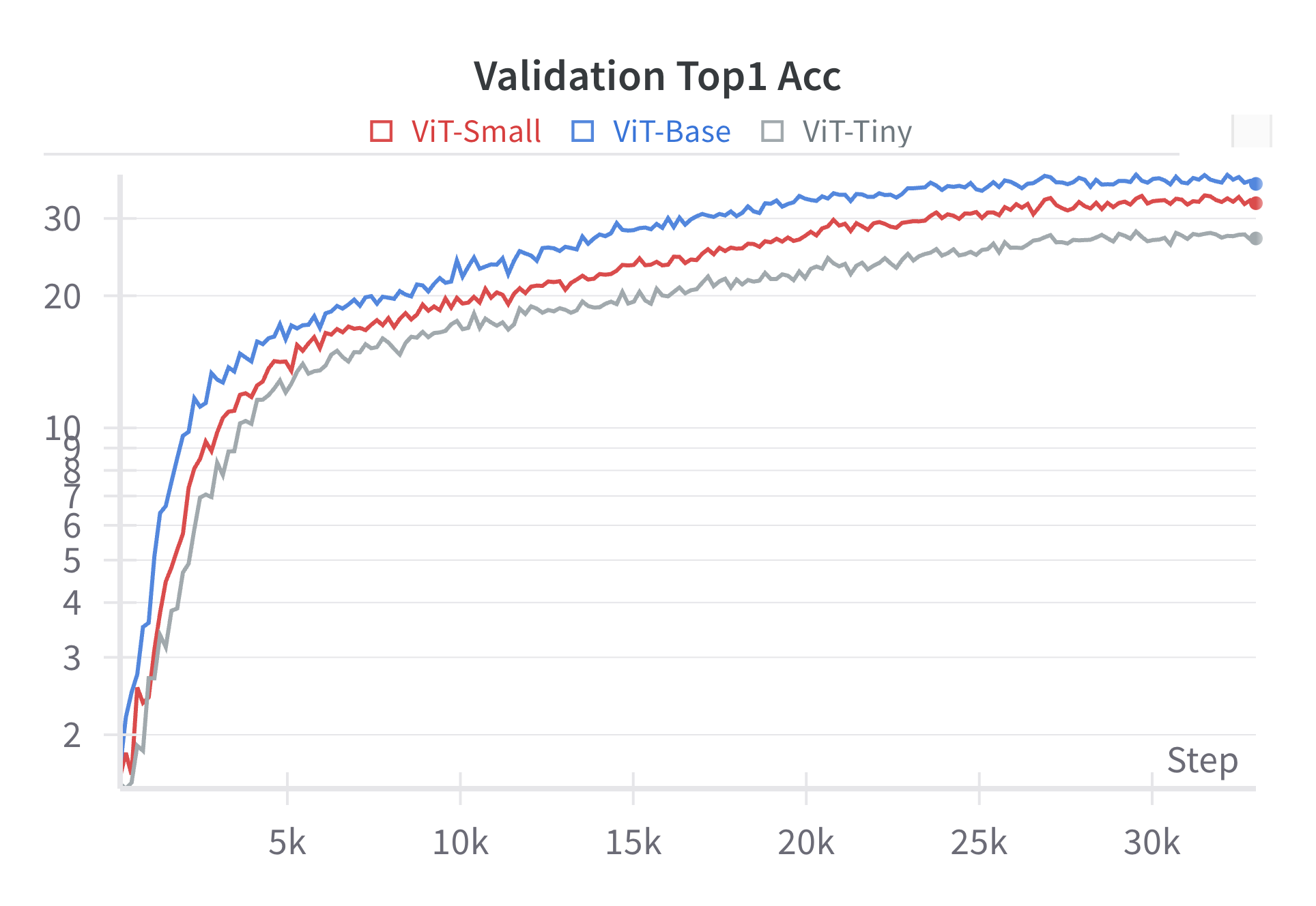}
        \label{fig:vit_vs_valtop1acc}
    \end{subfigure}
    \caption{While we find that the model scales on the dataset, the test set performance does not align with the validation set performance. One potential cause of this is distribution shift: the validation set uses pseudo-labels generated by GPT-4V, while the test set is human-labeled. }
    \label{fig:appendix_scale}
    \end{figure}
}

\def\figAppendixContact#1{
    \begin{figure}[#1]
    \centering
    \begin{subfigure}[b]{0.45\textwidth}
        \centering
        \includegraphics[width=\textwidth]{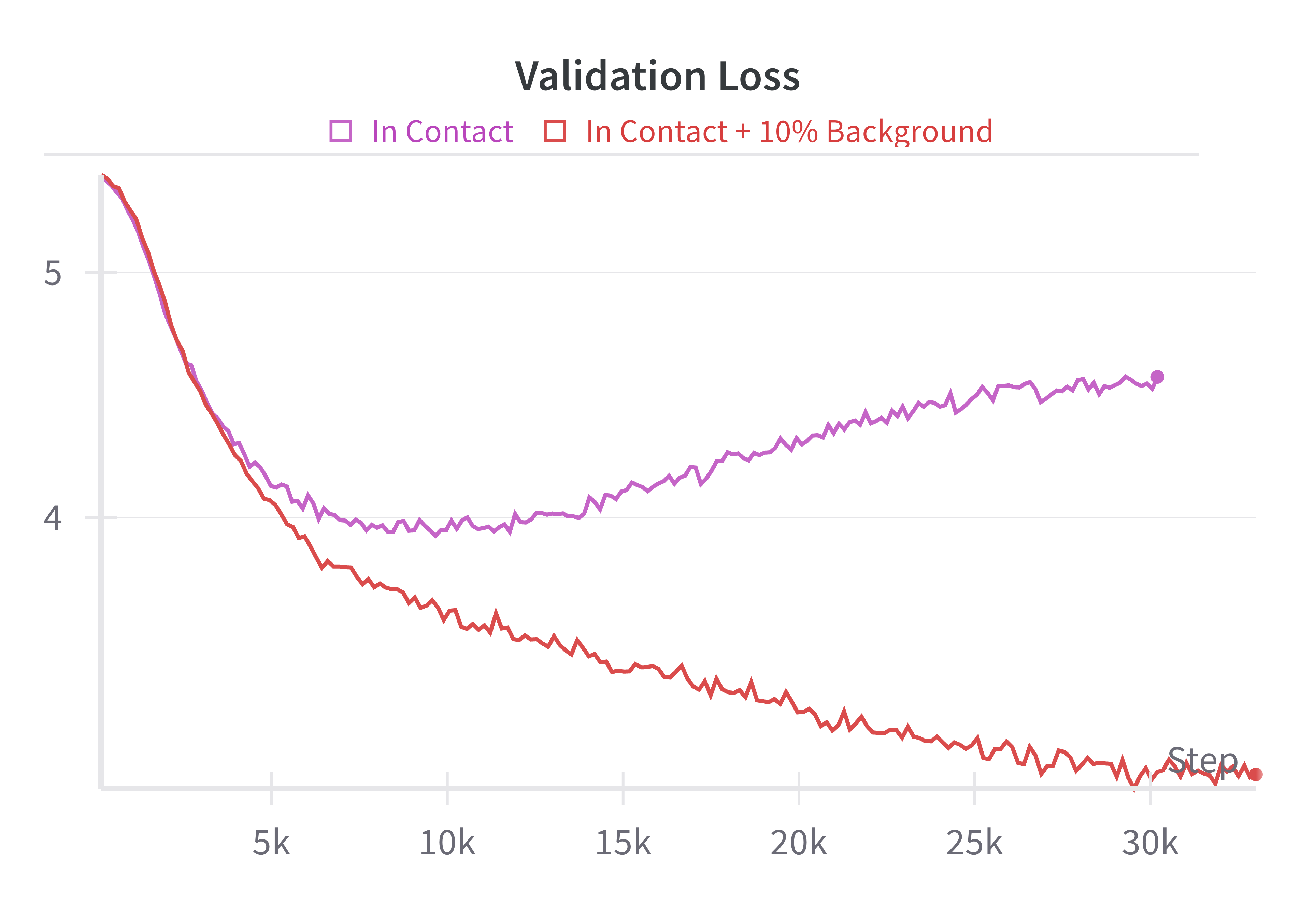}
        \label{fig:bv_vs_valloss}
    \end{subfigure}
    \hfill 
    \begin{subfigure}[b]{0.45\textwidth}
        \centering
        \includegraphics[width=\textwidth]{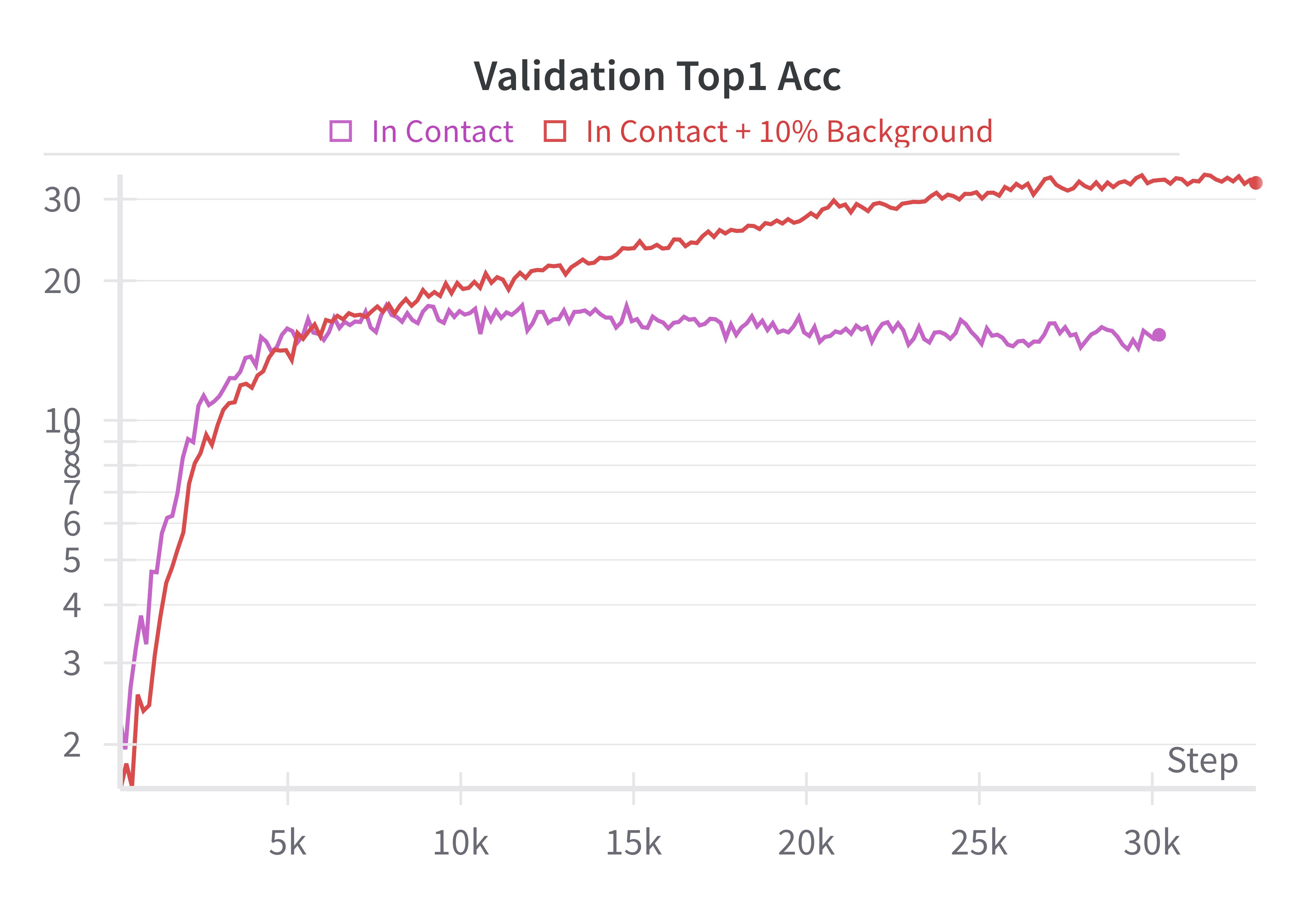}
        \label{fig:bg_vs_valtop1acc}
    \end{subfigure}
    \caption{Overfitting is significant when all data is in contact. When 10\% not in contact data is added, the overfitting issue is addressed.}
    \label{fig:appendix_contact}
    \end{figure}
}

\def\figSensorHolder#1{
    \begin{figure}[#1]
        \centering
        \includegraphics[width=1.0\linewidth]{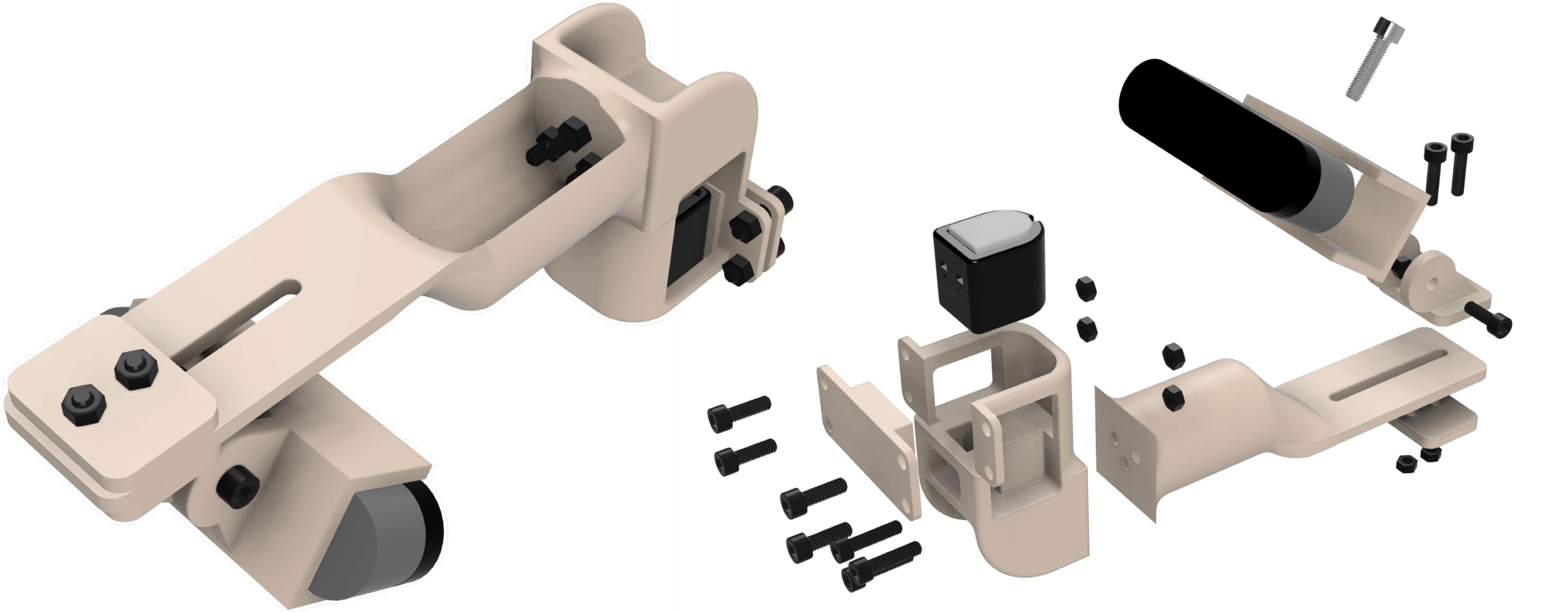}
        \caption{Alternative perspectives of the sensor holder CAD model: face-down view (left) and exploded view (right).}
        \label{fig:hardware}
    \end{figure}
}

\def\figWordDistribution#1{
    \begin{figure*}[#1]
        \centering
        \includegraphics[width=1.0\linewidth]{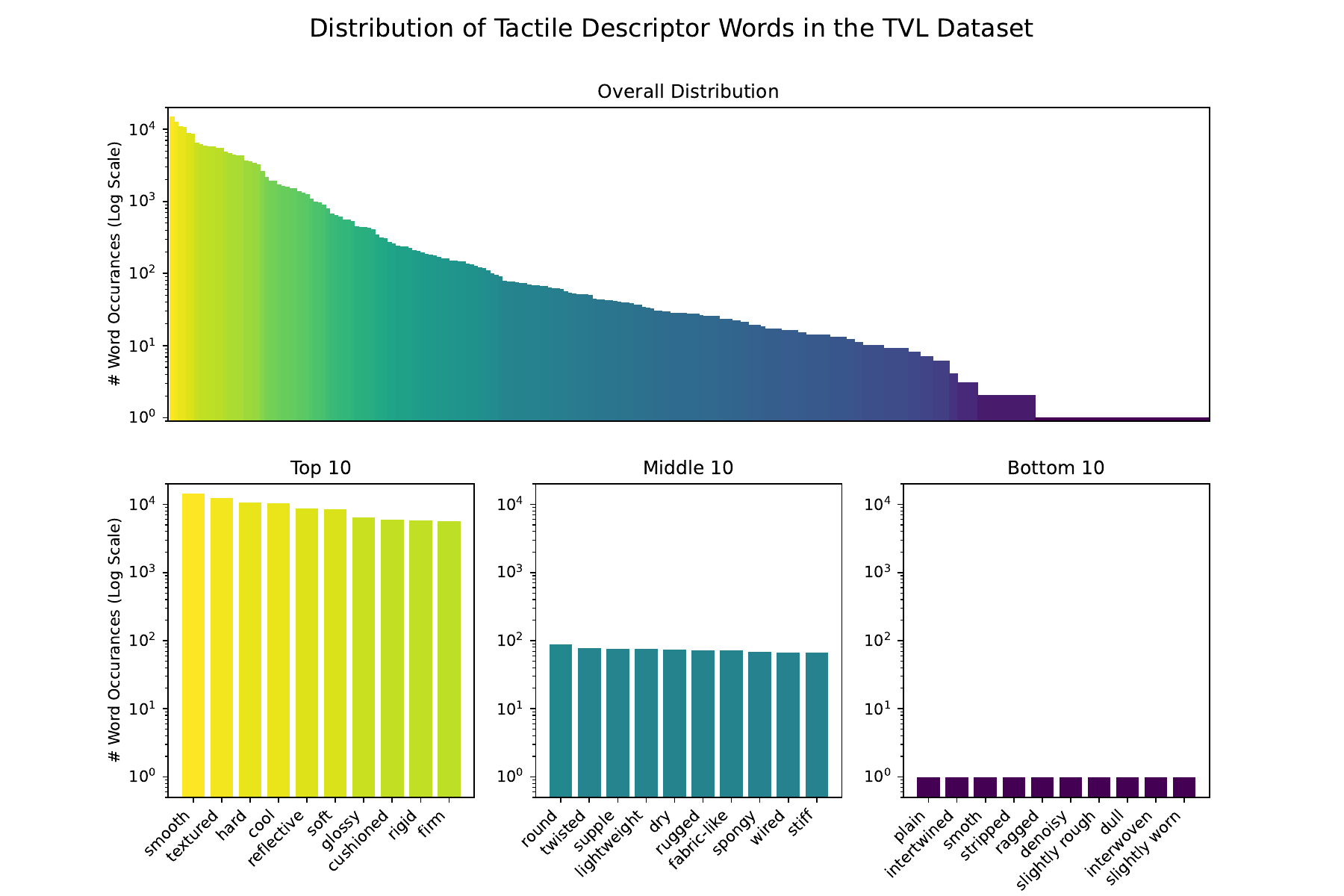}
        \vspace*{-0.1in}
        \caption{\textbf{Distribution of Words in the \algabbr Dataset:} The \algabbr dataset contains 254 unique tactile descriptors, ranging from common tactile descriptions (smooth, hard, firm) to unusual and optical descriptors. These less-common adjectives include a small fraction of misspellings and non-tactile descriptors which were generated by the VLM. The long-right-tailed distribution common in image classification~\cite {wang2020long} presents a challenge for learning predictors on tactile-semantic data as well.}
        \label{fig:distribution}
        \vspace*{0in}
    \end{figure*}
}
\section{Additional Results}
\label{sec:appendix:results}
\subsection{Performance Per Dataset}
\label{ssec:acc_dataset}
In this section, we show a fine-grained breakdown of \cref{tab:evaluation} of model performance on the TVU benchmark by showing the results per subset of the dataset. The performance of the models on the SSVTP subset is listed in \cref{tab:appendix_ssvtp_gen} and the performance on the HCT subset is listed in \cref{tab:appendix_hct_gen}. Results suggest that GPT-4V performs better on SSVTP, which is collected in a lab setting, than HCT, which is collected ``in-the-wild".

A model that is trained with a large sample of only GPT-4V labels should achieve the same performance as GPT-4V. Our results in \cref{tab:appendix_hct_gen} suggest that training on a small dataset of human-labeled vision-touch \textbf{improves} the model's tactile-visual understanding. This difference is statistically significant at $\alpha=0.05$. 

\tabPerDatasetSSVTP{h}
\tabPerDatasetHCT{h}

\subsection{Open Vocabulary Tactile Classification Full Result}
\label{ssec:appendix_classification}
We present the result presented in \cref{tab:encoder} in \cref{tbl:enc_tac_text} and \cref{tbl:enc_tacvis} at different cosine similarity threshold for synonyms. We find that while ViT-Small performs well on the SSVTP subset of the dataset, ViT-Tiny outperforms its larger counterparts (ViT-Small and ViT-Base) on the tactile-text classification task. However, for tactile-vision classification (\cref{tbl:enc_tacvis}), ViT-Base performs outperforms the smaller models. More insights are detailed in \cref{sec:appendix:overfit}. 

\tabenctactext{h}
\tabenctacvis{h}

\section{Training Details and Hyperparameters}
In this section, we offer more insights and details of the training process and the particular hyperparameters. 
\subsection{Overfitting to Pseudo-labels}
\label{sec:appendix:overfit}
A core obstacle with leveraging pseudo-labels generated by GPT-4V (gpt-4-vision-preview) is that the logits are not provided for us to build uncertain estimates for the generated labels, which is usually required for prior works in computer vision that leverages pseudo-labels for model prediction (\textit{e.g.} ~\citet{sohn2020fixmatch, lee2013pseudo, wang2022debiased}). This makes pseudo-labels noisy and challenging to fit for ViT-Small on the contact only dataset, even when 4K human labels are introduced (see \cref{fig:appendix_contact}).
\figAppendixContact{h}

In \ref{ssec:tac_enc}, we address this problem by letting 10\% of the data be in contact. We sample 10\% of the data uniformly at random without replacement at the start of the training. This prevents the model from overfitting on all three model sizes: (ViT-Tiny, ViT-Small, and ViT-Base). However, since the test set is all labeled by human annotators, the distribution shift leads to worse tactile-image, and tactile-language classification performance (observed in \cref{tab:encoder}). As an ablation study, we also finetuned the ViT-Small trained only on in-contact data for tactile language generation. The test set performance is 4.81, only very marginally lower than that obtained by the ViT-Small trained with not-in-contact data (4.89). Future works can look into how to scale with noisy inputs or leverage existing works on learning from a teacher model that does not give uncertain estimates. 
\figAppendixScaling{h}

\subsection{Ablation: Background Subtraction}
\label{sec:appendix_bgs}
While we find that naively performing contrastive learning amongst tactile, vision, and language works for zero-shot classification, to further facilitate generalization across different tactile sensors used in data collection, a solution is to leverage the still background of tactile sensors (\textit{i.e.} the readings from the sensor when it is not in contact). We preprocess the tactile observation by performing background subtraction, and normalize the input observations based on the post-processed dataset statistics. Empirically, we find that this method, when used jointly with not-in-contact data, improves classification accuracy and the downstream TVL-LLaMA's performance (\cref{tab:background}).
\tabBGS{h}

\subsection{Ablation: (Zero-shot) Single Modality For Generation (Out of Distribution)}
Because we naively average the tactile latent and the image latent during the training of \algabbr-LLaMA, as a zero-shot experiment to see consistency between vision and tactile embeddings, we can at \textit{test} time arbitrarily drop one of the vision or tactile modalities. We report the results in \cref{tab:ood_zeroshot}. While a larger encoder may be more expressive, we find that a larger tactile encoder results in worse zero-shot performance in this experimental setting, which aligns with \cref{tab:ablation-architecture}. Interestingly, background subtraction (in \cref{sec:appendix_bgs}) improves the zero-shot performance on tactile. 
\tabZeroShot{h}

\subsection{Preprocessing}
The tactile observation is first zero-padded to have equal width and height, optionally background subtracted, normalized by the calculated data statistics, and resized the inputs to 224x224. The key differences with SSVTP are 1) the input is resized to 128x128, and 2) SSVTP does not perform normalization or background subtraction. The image observation follows the same center cropping procedure as SSVTP on the SSVTP dataset. On \dataabbr, instead of the center crop, we start the crop from the top of the image but maintain the crop size. Note that this procedure is kept consistent when generating pseudo-labels from GPT-4V. Different from SSVTP, we use the statistics provided by OpenCLIP to normalize the post-crop observations. The specific statistics are provided in \cref{tab:appendix_tacstat} and \cref{tab:appendix_rgbstat}.
\tabtactilestat{h}
\tabrgbstat{h}
\subsection{\algabbr Tactile Encoder Hyperparameters}
All of ViT-Tiny, ViT-Small, and ViT-Base share the same hyperparameters (see \cref{tbl:enc_hyper}). All experiments are run on a single NVIDIA A100 GPU.
\tabenchyper{h}

\subsection{\algabbr-LLaMA Hyperparameters}
We follow the hyperparameter setup in ImageBind-LLM~\cite{han2023imagebindllm}. Since the original experiments were conducted on 8 NVIDIA A100 GPUs, we use gradient accumulation of 2 for both pre-training and finetuning the model to fit the model on 4 NVIDIA A100 GPUs so that the batch size is maintained. We use the same data augmentation as in the encoder pretraining (\cref{tbl:enc_hyper}).

\section{Dataset}
\subsection{Hardware}

\figSensorHolder{h}

We design and 3D print a set of handheld, low-cost data collection devices for human subjects to carry around and collect data. As shown in Fig.~\ref{fig:hardware}, the hardware consists of a DIGIT tactile sensor and a Logitech BRIO camera, which are connected via USB to a portable computing device, such as a laptop. The angle and distance between the tactile sensor and the camera are adjustable, allowing the user to collect data from a variety of viewing angles and ranges. To ensure the utility of our dataset for multimodal training, we always set the relative positions such that the tactile sensor and its point of contact with the object of interest are in view of the camera during each trajectory. The handle design was conceptualized in Autodesk Fusion 360 and printed on a Bambu Lab P1P 3D FDM printer. CAD files will be open-sourced.

\subsection{List of Prompts for Tactile Language Generation}
\label{sec:appendix:vqa_prompts}
When finetuning our language model for tactile language generation, we formulate it as a visual instruction tuning problem~\cite{liu2023llava}. We randomly select from the following set of semantically similar prompts as the question and treat the set of human labels as the answer. This serves to increase the diversity of data seen during training.

\begin{lstlisting}[numbers=none]
This image gives tactile feelings of 
This image evokes a sense of  
This visual representation imparts a tactile sensation of 
This picture conveys a touchable quality of 
This image communicates a palpable feeling of 
This graphic suggests a tactile experience of 
This artwork manifests a tangible sensation of 
This visual elicits a haptic impression of 
This depiction gives rise to a tactile perception of 
This illustration induces a touch-sensitive feeling of 
This photo brings forth a tactile awareness of 
This image arouses a tactile familiarity of 
This snapshot renders a tactile essence of 
This visual stimulates a touch-based sensation of 
This portrayal invokes a tactile resonance of 
This image delivers a touch-oriented impression of 
This visual medium offers a tactile nuance of 
This rendering provides a tactile sense of 
This image yields a touch-felt experience of 
This composition reveals a tactile characteristic of 
This picture bestows a tactile attribute of 
This image imparts a sense of tactile 
This visual stimulates tactile sensations of 
This artwork hints at a tactile experience of 
This photo embodies a tactile quality of 
This depiction resonates with tactile feelings of 
This snapshot conveys tactile impressions of 
This illustration suggests a tactile nature of 
This rendering evokes tactile attributes of 
This graphic communicates a tactile essence of 
This visual piece reveals tactile characteristics of 
This image portrays tactile elements of 
This picture brings to mind tactile aspects of 
This visual representation offers tactile nuances of 
This composition provides tactile insights into 
This visual art form captures tactile features of 
This image projects tactile properties of 
This visual work hints at tactile textures of 
This image introduces tactile dimensions of 
This visual scene manifests tactile facets of 
This image presents tactile qualities of 
This image elucidates tactile attributes of 
\end{lstlisting}

\subsection{Distribution of Vocabulary Words}
\label{sec:appendix:distr}
The list and counts of human labels and pseudo-labels in the \algabbr dataset are reproduced here in dictionary format (note that all typos are carried over from the dataset). A visual representation is provided in \cref{fig:distribution}.

\figWordDistribution{htb}

{'smooth': 14577, 'textured': 12443, 'hard': 10758, 'cool': 10433, 'reflective': 8643, 'soft': 8415, 'glossy': 6416, 'cushioned': 6011, 'rigid': 5799, 'firm': 5659, 'sleek': 5628, 'uneven': 5379, 'flat': 5343, 'fibrous': 4825, 'plush': 4534, '': 4363, 'matte': 4230, 'polished': 4203, 'flexible': 3553, 'grainy': 3513, 'solid': 3337, 'warm': 3227, 'woven': 2559, 'fabric': 2124, 'yielding': 1908, 'rough': 1889, 'slippery': 1683, 'slick': 1587, 'rubbery': 1553, 'coarse': 1504, 'lined': 1480, 'durable': 1362, 'pliable': 1281, 'curved': 1240, 'bumpy': 1076, 'metallic': 970, 'patterned': 949, 'cloth-like': 889, 'resilient': 785, 'abrasive': 668, 'plastic': 631, 'ridged': 599, 'gritty': 551, 'deformable': 544, 'compressible': 517, 'synthetic': 444, 'fuzzy': 434, 'varnished': 430, 'dimpled': 423, 'wooden': 399, 'thin': 337, 'irregular': 311, 'splotchy': 301, 'even': 267, 'uniform': 257, 'perforated': 239, 'granular': 234, 'indistinct': 230, 'plastic-like': 220, 'grooved': 204, 'paper-like': 203, 'blurred': 191, 'sewn': 183, 'elastic': 179, 'contoured': 173, 'shiny': 165, 'blurry': 159, 'level': 159, 'taut': 149, 'grid-like': 149, 'creased': 145, 'porous': 145, 'grippy': 135, 'cushiony': 132, 'speckled': 126, 'leather-like': 120, 'grained': 116, 'knitted': 107, 'padded': 99, 'worn': 94, 'round': 89, 'twisted': 77, 'supple': 76, 'lightweight': 76, 'dry': 73, 'rugged': 72, 'fabric-like': 72, 'spongy': 69, 'wired': 67, 'stiff': 67, 'unclear': 66, 'indented': 66, 'dense': 62, 'dark': 61, 'iridescent': 61, 'undefined': 59, 'knobby': 55, 'grid-patterned': 53, 'layered': 52, 'resonant': 51, 'fluffy': 50, 'translucent': 50, 'soft-focus': 49, 'absorbent': 44, 'slightly textured': 43, 'leathery': 43, 'obscured': 42, 'cylindrical': 42, 'wrinkly': 41, 'unfocused': 40, 'ribbed': 39, 'rippled': 39, 'thick': 38, 'sturdy': 36, 'striated': 36, 'hairy': 34, 'hazy': 33, 'embroidered': 32, 'raised': 30, 'cottony': 30, 'colorful': 29, 'slightly compressible': 29, 'straight': 28, 'silky': 28, 'braided': 28, 'straight-edged': 28, 'overexposed': 27, 'angular': 27, 'ethereal': 27, 'glowing': 26, 'lettered': 25, 'tough': 25, 'edged': 25, 'rounded': 25, 'transparent': 23, 'smeared': 23, 'carpeted': 23, 'stretchy': 22, 'slightly squishy': 22, 'fleshy': 21, 'ceramic': 21, 'engraved': 19, 'opaque': 19, 'clothlike': 19, 'bright': 18, 'folded': 17, 'striped': 17, 'embossed': 17, 'brushed': 17, 'mesh': 16, 'stable': 16, 'bendable': 16, 'slightly bendable': 16, 'frayed': 15, 'printed': 15, 'vague': 14, 'cardboard': 14, 'clickable': 14, 'organic': 14, 'delicate': 14, 'undulating': 14, 'clear': 13, 'stringy': 13, 'clicky': 13, 'smooth edges': 13, 'sticky': 12, 'out-of-focus': 12, 'lace': 11, 'brittle': 11, 'regular': 10, 'open': 10, 'continuous': 10, 'muted': 10, 'slightly abrasive': 10, 'malleable': 9, 'incised': 9, 'motion-blurred': 9, 'slightly warm': 9, 'intricate': 9, 'obscure': 9, 'laced': 8, 'slightly curved': 8, 'compliant': 8, 'metal': 7, 'sewed': 7, 'pressed': 7, 'flimsy': 6, 'sandy': 6, 'insulated': 6, 'convex': 6, 'sharp': 4, 'crinkled': 4, 'springy': 3, 'complex': 3, 'grainy fabric': 3, 'line': 3, 'slightly gritty': 3, 'consistent': 2, 'loose': 2, 'paper': 2, 'fraying': 2, 'lustrous': 2, 'spotty': 2, 'light': 2, 'bristly': 2, 'woolen': 2, 'wrinkled': 2, 'griany': 2, 'precise': 2, 'non-glossy': 2, 'wavy': 2, 'lacey': 1, 'meshed': 1, 'imprinted': 1, 'flat smooth': 1, 'sewn fabric': 1, 'shadow': 1, 'bendy': 1, 'rigit': 1, 'jagged': 1, 'flash': 1, 'frabric': 1, 'patterened': 1, 'floor': 1, 'flawless': 1, 'long': 1, 'spolotchy': 1, 'granulated': 1, 'cloth': 1, 'thready': 1, 'patterend': 1, 'smooth fabric': 1, 'deformalbe': 1, 'smmoth': 1, 'wirey': 1, 'fabric granular': 1, 'graint': 1, 'lined sewn': 1, 'smotth': 1, 'wiry': 1, 'torn': 1, 'vauge': 1, 'facrib': 1, 'gariny': 1, 'plain': 1, 'intertwined': 1, 'smoth': 1, 'stripped': 1, 'ragged': 1, 'denoisy': 1, 'slightly rough': 1, 'dull': 1, 'interwoven': 1, 'slightly worn': 1}

\subsection{Prompting for Psuedo-Label Generation}
\label{sec:appendix:pseudo}
We use the following prompt with GPT-4V in order to label the images with tactile descriptions:

\begin{lstlisting}
Surface Type: [Specify the surface type, e.g., "metal," "fabric"]
Images: The first image is from a camera observing the tactile sensor (shiny, near the top of the image) and the surface. The second image is a cropped version of the first image that focuses on the contact patch. 
Example: For a smooth and cold surface, the description might be "slick, chilly, hard, unyielding, glossy."
Task: Based on these images, describe the possible tactile feelings of the contact patch using sensory adjectives. Limit your response up to five adjectives, separated by commas.
\end{lstlisting}

\subsection{Prompting GPT-4 for Evaluation}
\label{sec:appendix:evaluation_prompt}
We use the following prompt for TVL Benchmark:

\begin{lstlisting}
[User Question]: {prompt}
[Assistant Response]: {assistant_response}
[Correct Response]: {correct_response}

We would like to request your feedback on the performance of an AI assistant in response to the user question displayed above. 
The user asks the question on observing an image. The assistant's response is followed by the correct response.

Please evaluate the assistant's response based on how closely it matches the correct response which describes tactile feelings. Please compare only the semantics of the answers. DO NOT consider grammatical errors in scoring the assistant. The assistant receives an overall score on a scale of 1 to 10, where a higher score indicates better overall performance.

Please first output a single line containing only one value indicating the score for the assistant. 

In the subsequent line, please provide a comprehensive explanation of your evaluation, avoiding any potential bias.
\end{lstlisting}

\section{Generation Examples}
We provide a few positive and negative samples of image-tactile pairs from our dataset and the language descriptions generated for them by our various baseline models.

\begin{figure}[h]
    \centering
    \includegraphics[width=1.0\linewidth]{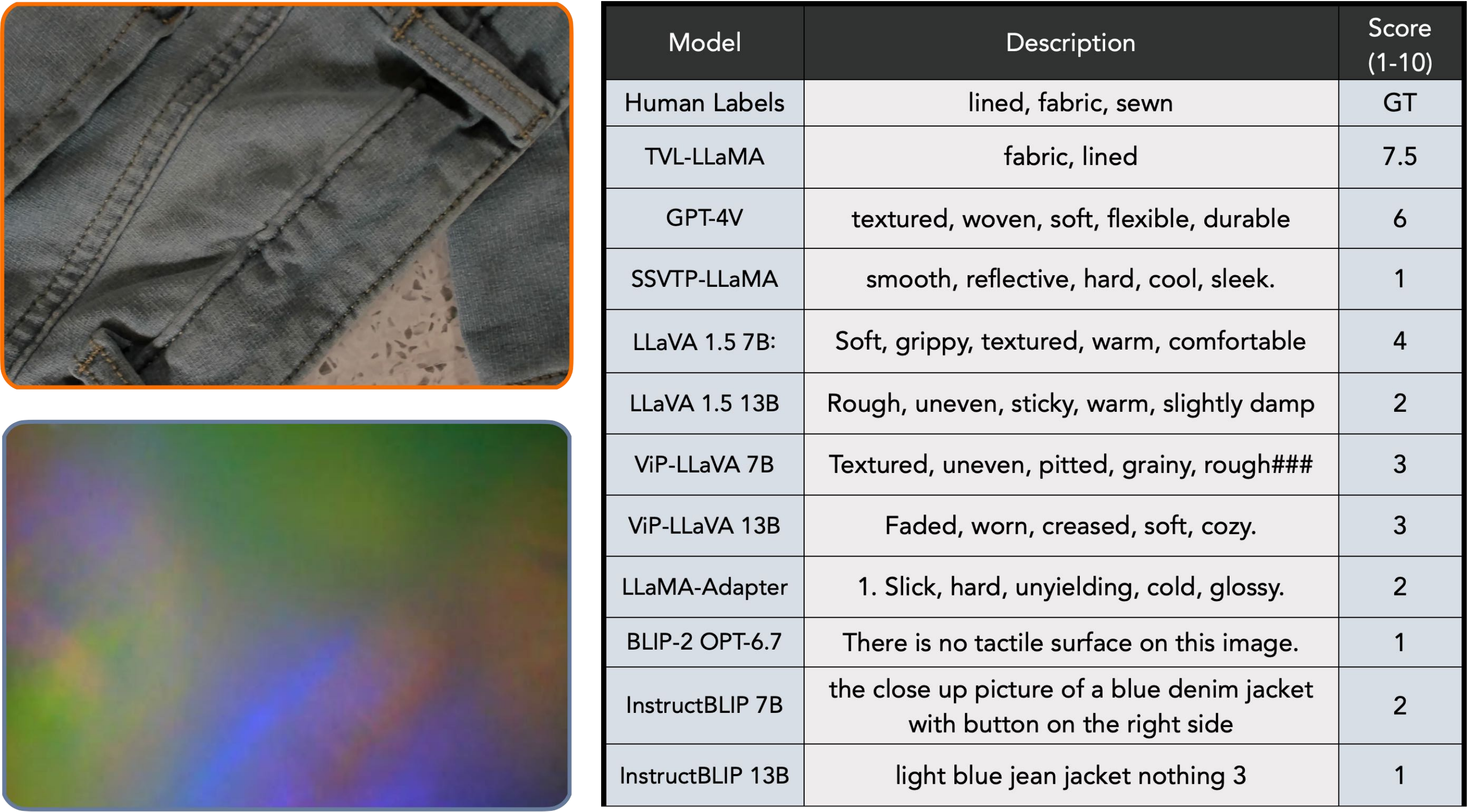}
    \label{fig:enter-label}
\end{figure}
\begin{figure}[h]
    \centering
    \includegraphics[width=1.0\linewidth]{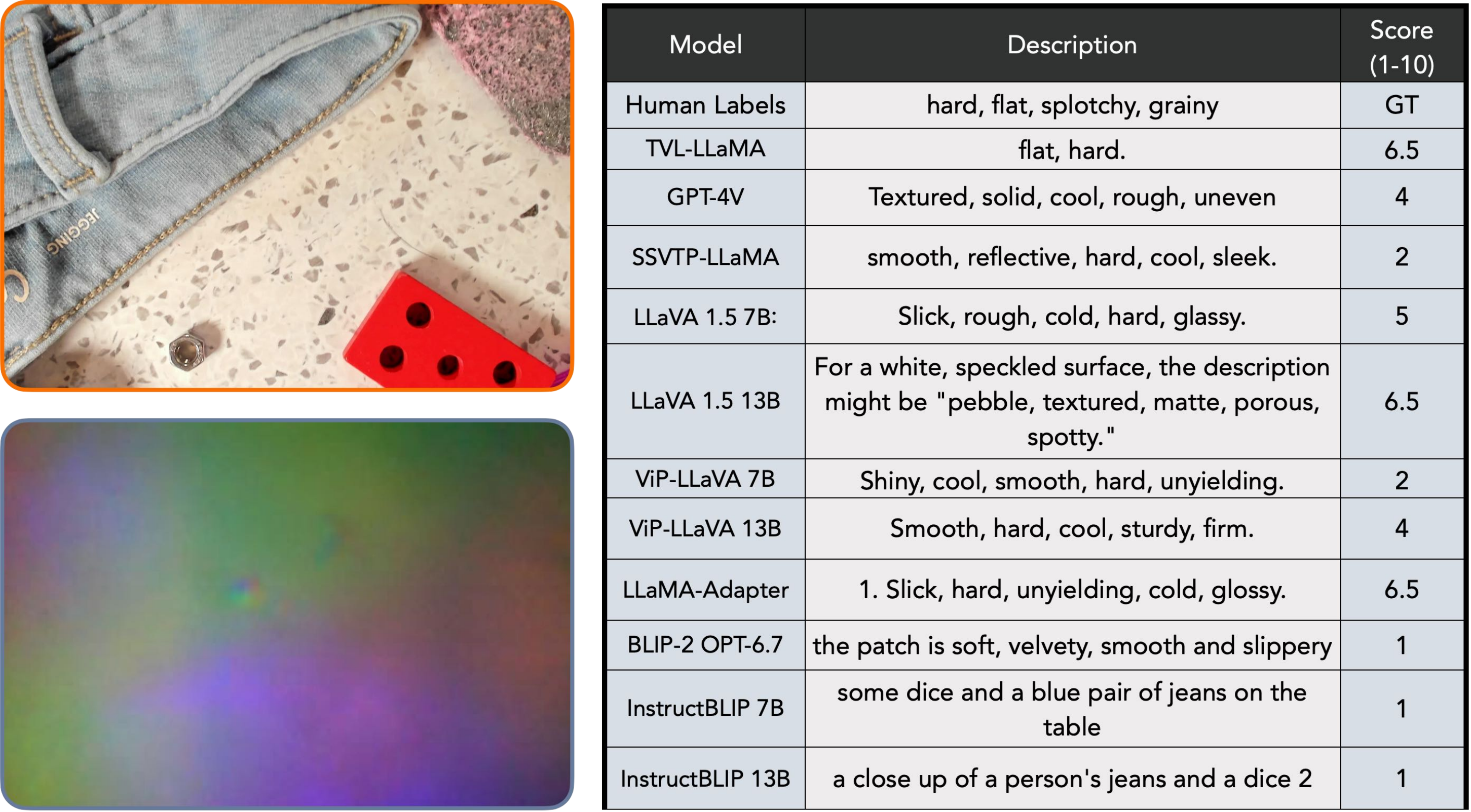}
    \label{fig:enter-label}
\end{figure}
\begin{figure}[h]
    \centering
    \includegraphics[width=1.0\linewidth]{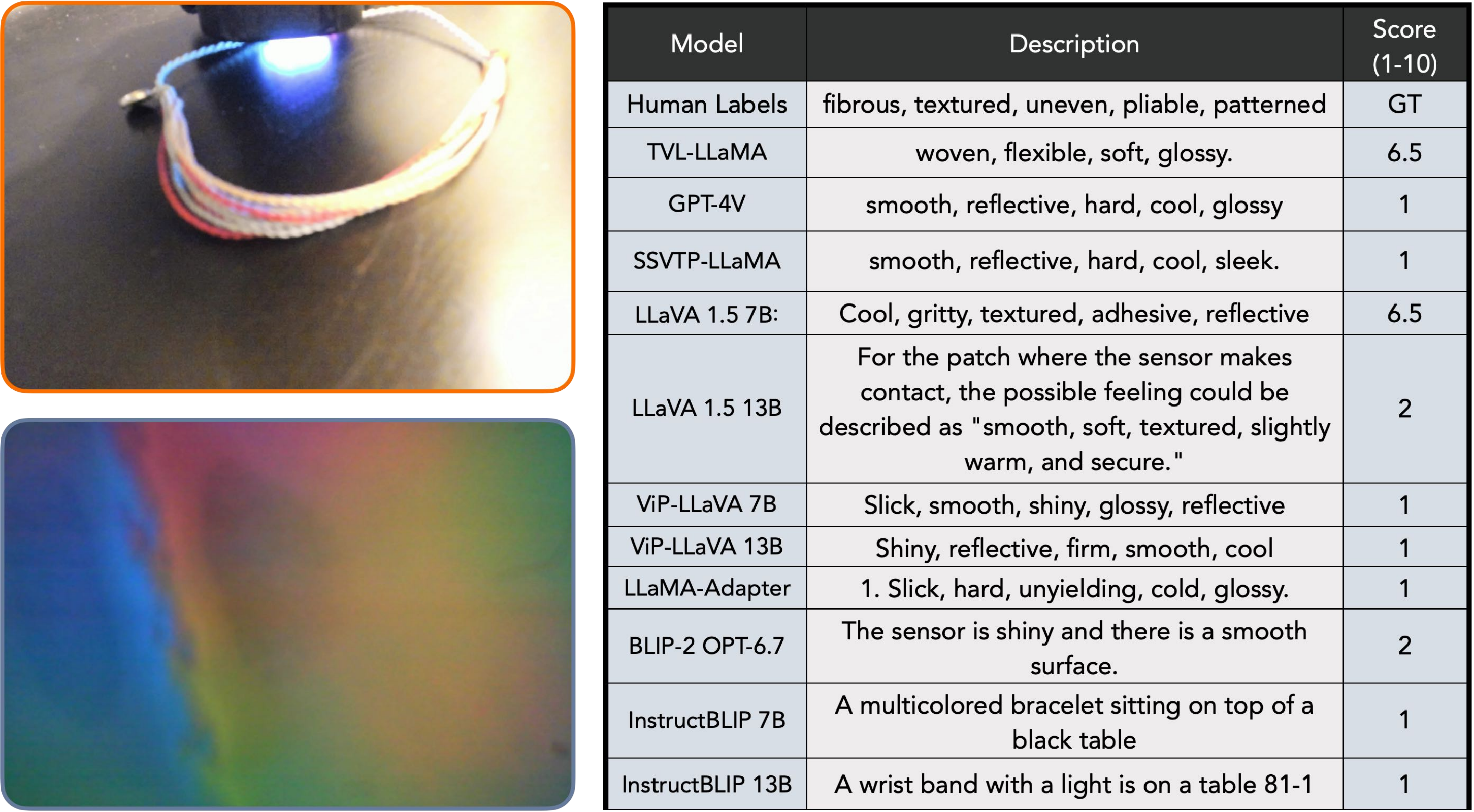}
    \label{fig:enter-label}
\end{figure}
\begin{figure}[h]
    \centering
    \includegraphics[width=1.0\linewidth]{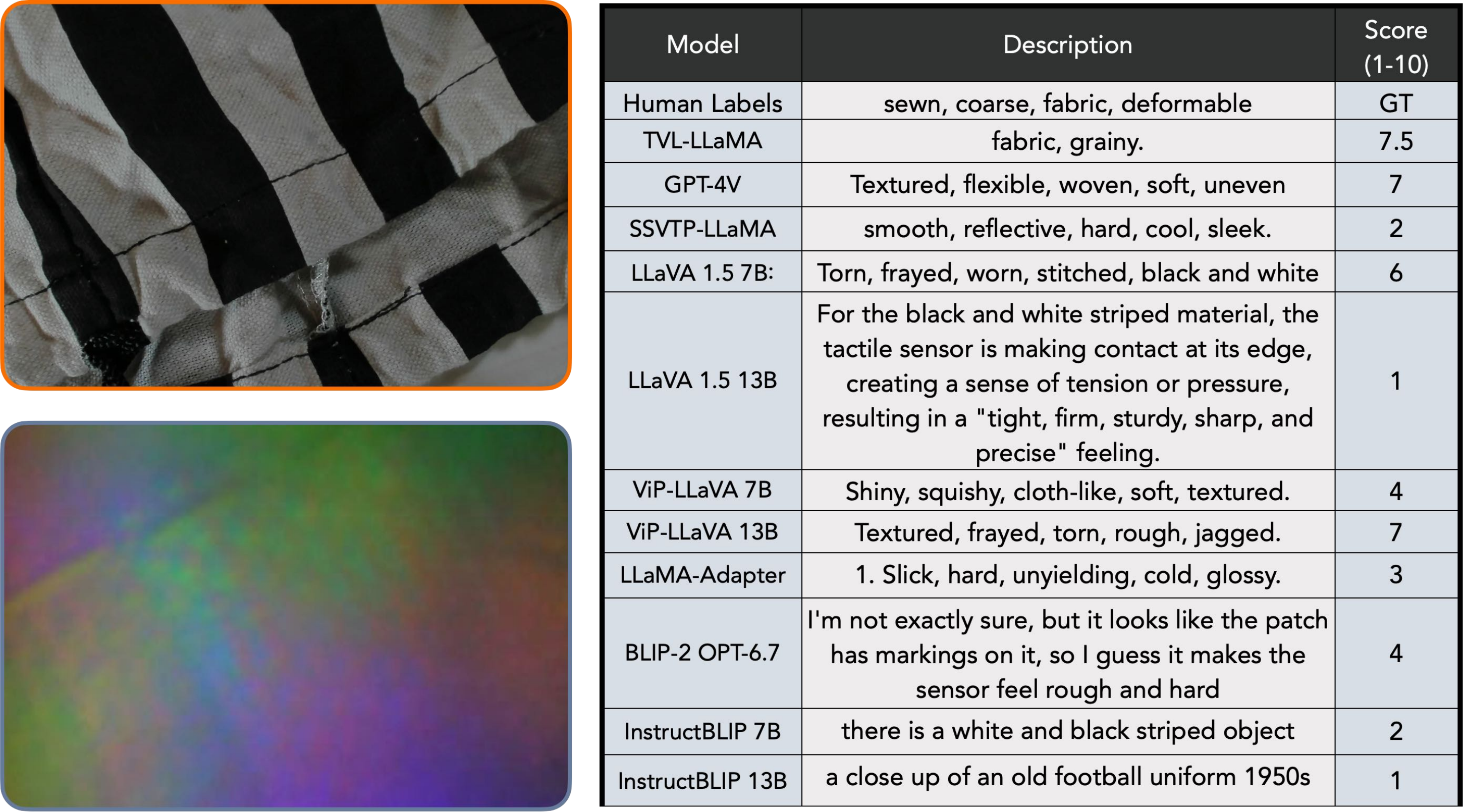}
    \label{fig:enter-label}
\end{figure}
\begin{figure}[h]
    \centering
    \includegraphics[width=1.0\linewidth]{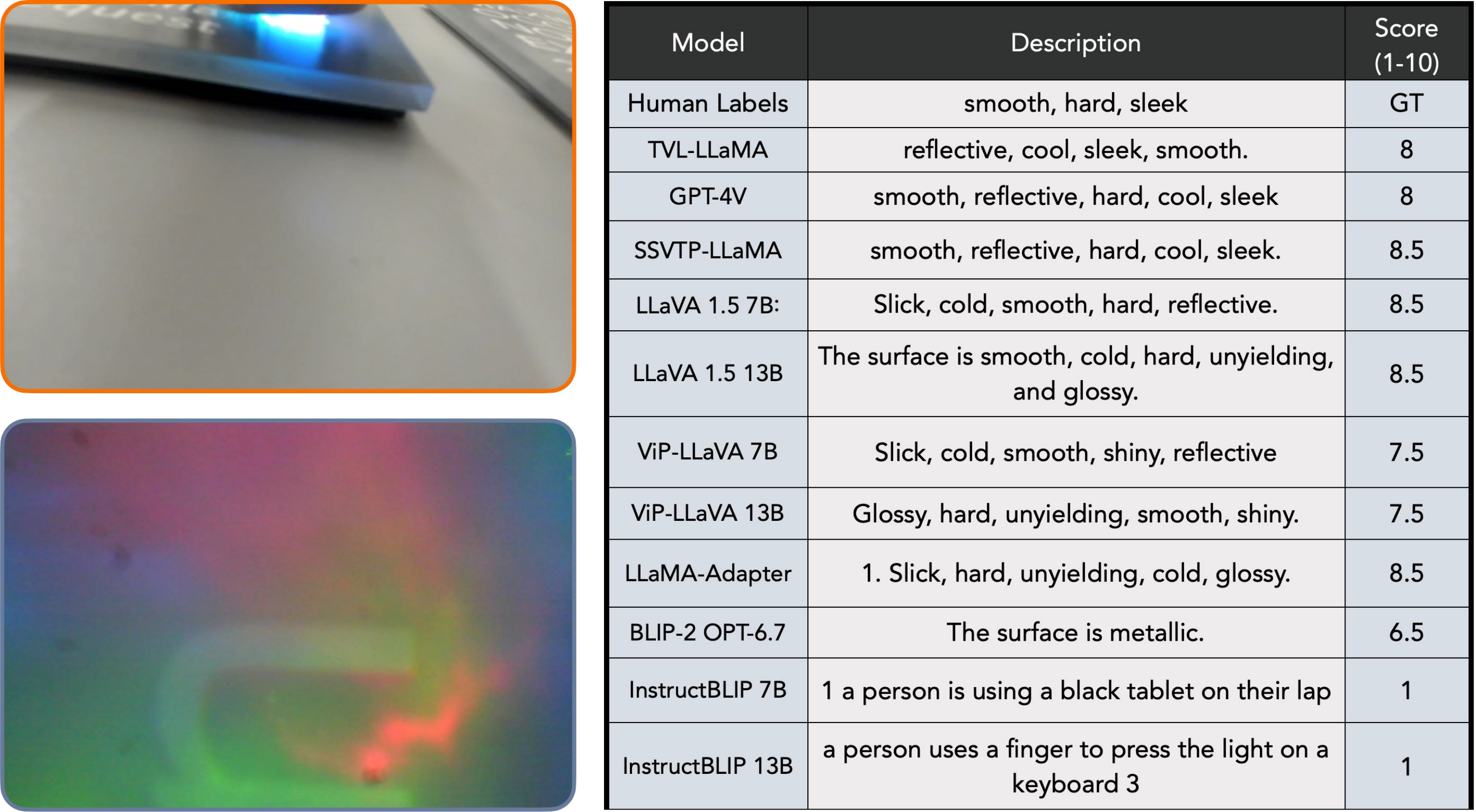}
    \label{fig:enter-label}
\end{figure}
\begin{figure}[h]
    \centering
    \includegraphics[width=1\linewidth]{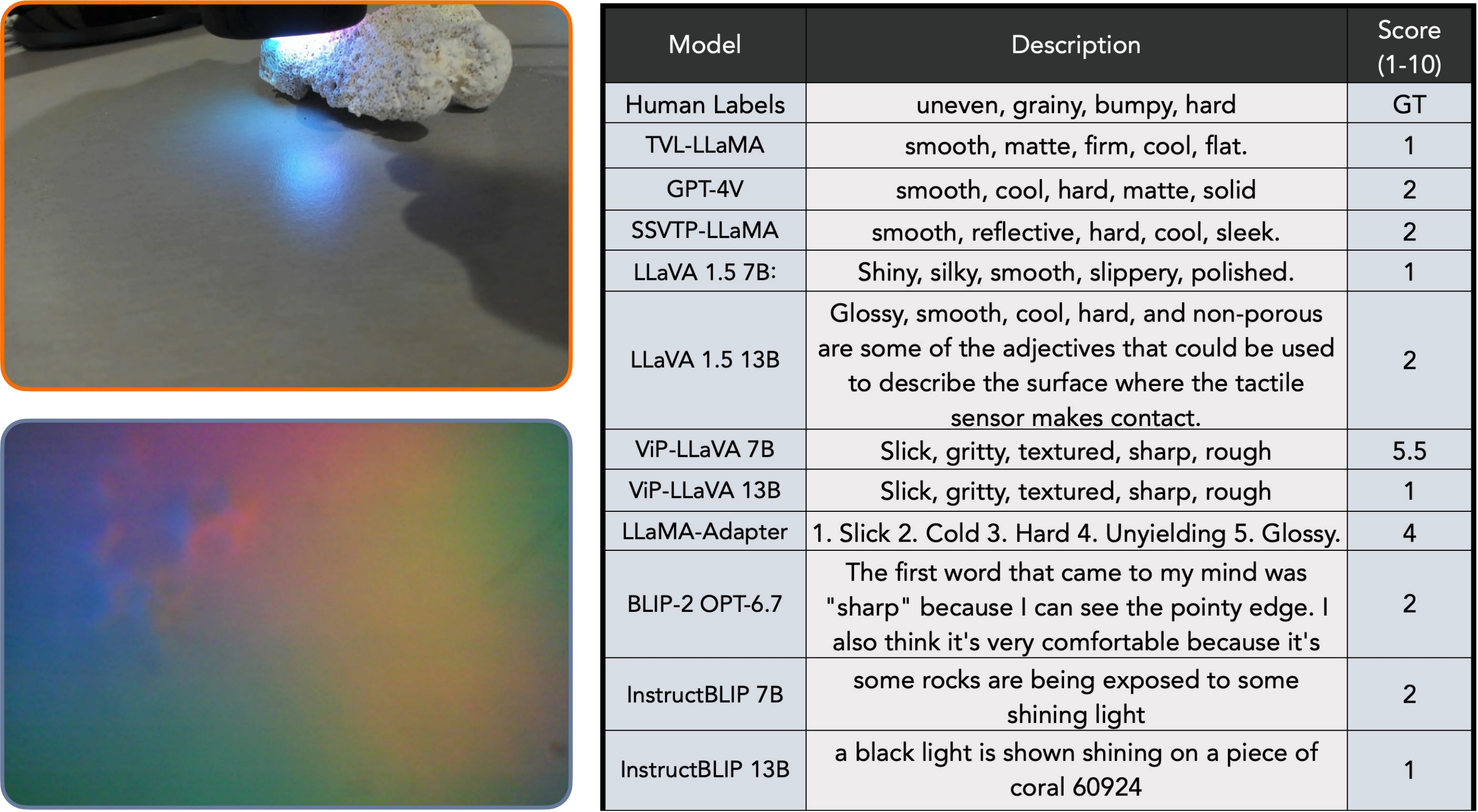}
    
    \label{fig:enter-label}
\end{figure}
\begin{figure}[h]
    \centering
    \includegraphics[width=1\linewidth]{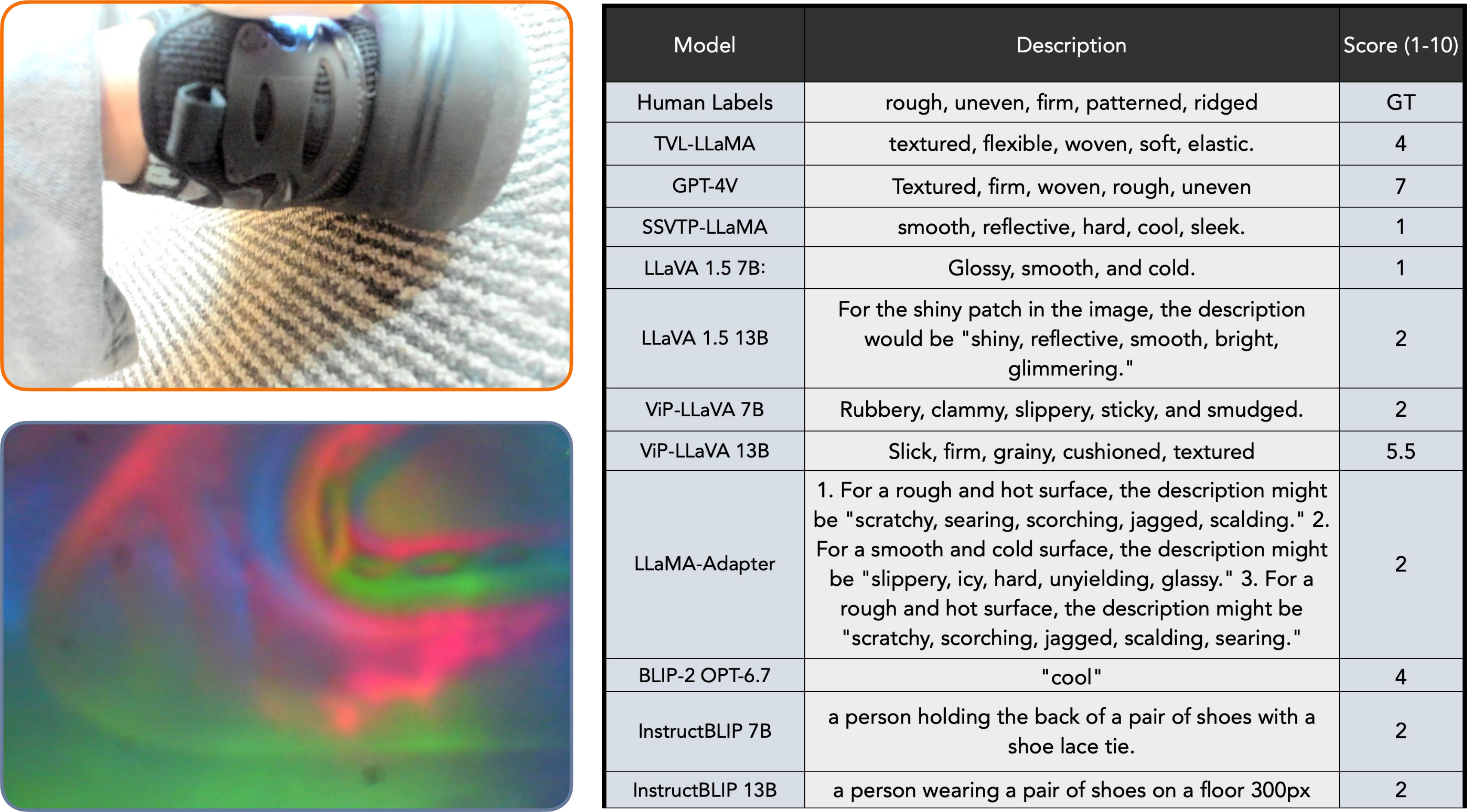}
    
    \label{fig:enter-label}
\end{figure}

\end{document}